\documentclass[10pt,twocolumn,letterpaper]{article}

\usepackage{iccv}
\usepackage{times}
\usepackage{epsfig}
\usepackage{graphicx}
\usepackage{amsmath}
\usepackage{amssymb}
\usepackage{setspace}
\usepackage[accsupp]{axessibility}

\usepackage{graphicx}
\usepackage{comment}
\usepackage{amsmath,amssymb,amstext} %

\usepackage{gensymb}
\belowdisplayskip \abovedisplayskip

\newlength{\sectionReduceTop}
\newlength{\sectionReduceBot}
\newlength{\subsectionReduceTop}
\newlength{\subsectionReduceBot}
\newlength{\abstractReduceTop}
\newlength{\abstractReduceBot}
\newlength{\captionReduceTop}
\newlength{\captionReduceBot}
\newlength{\subsubsectionReduceTop}
\newlength{\subsubsectionReduceBot}

\newlength{\eqnReduceTop}
\newlength{\eqnReduceBot}

\newlength{\horSkip}
\newlength{\verSkip}

\newlength{\figureHeight}
\usepackage{relsize}
\usepackage[T1]{fontenc}
\usepackage[scaled=0.90]{inconsolata}
\usepackage[latin1]{inputenc}
\usepackage[english]{babel}

\usepackage{epsfig}
\usepackage{graphicx}
\usepackage{wrapfig}
\usepackage[belowskip=2pt,labelfont=bf,aboveskip=0pt,font=small,labelsep=period]{caption}
\usepackage{array}

\usepackage[belowskip=0pt,aboveskip=0pt,font=small]{subcaption}
\usepackage{longtable}
\usepackage{microtype}

\usepackage{textcomp}
\usepackage{stmaryrd}
\SetSymbolFont{stmry}{bold}{U}{stmry}{m}{n}
\usepackage{upgreek}
\usepackage{bm}
\usepackage{cases}
\usepackage{mathtools}
\usepackage{arydshln}
\usepackage{multirow}

\usepackage{color}
\usepackage[dvipsnames]{xcolor}
\definecolor{JalapenoRed}{RGB}{183,21,64}
\definecolor{Belize}{RGB}{41,128,185}
\definecolor{Amour}{RGB}{238,82,83}

\usepackage{algorithm}
\usepackage{algpseudocode}

\usepackage{multirow}
\usepackage{rotating}
\usepackage{booktabs}
\usepackage{etoolbox}
\newcounter{magicrownumbers}
\preto\tabular{\setcounter{magicrownumbers}{0}}
\newcommand\rownumber{\stepcounter{magicrownumbers}\arabic{magicrownumbers})\,}

\usepackage{enumitem}
\usepackage[olditem,oldenum]{paralist}

\usepackage{alltt}
\usepackage{listings}

\usepackage{url}
\usepackage{xspace}
\usepackage{comment}
\usepackage{afterpage}
\usepackage{pdfpages}
\usepackage{framed}
\usepackage{fancybox}
\usepackage{cuted}
\usepackage{caption}

\usepackage{quoting}
\quotingsetup{vskip=0pt}
\usepackage{epigraph}
\usepackage[normalem]{ulem}

\setdefaultleftmargin{.2cm}{.2cm}{}{}{}{}
\setlist{leftmargin=.2cm}

\usepackage{mysymbols}
\belowdisplayskip \abovedisplayskip

\usepackage[numbers,sort,compress]{natbib}

\usepackage[pagebackref=true,breaklinks=true,colorlinks,bookmarks=false,letterpaper=true,allcolors=JalapenoRed]{hyperref}

\iccvfinalcopy %

\def\iccvPaperID{7270} %

\ificcvfinal\fi

\begin{document}

\title{Auxiliary Tasks and Exploration Enable ObjectGoal Navigation}

\author{
    Joel Ye$^{1}$\thanks{Correspondence to \texttt{joel.ye@gatech.edu}}\quad
    Dhruv Batra$^{1,2}$\quad
    Abhishek Das$^{2}$\quad
    Erik Wijmans$^{1}$\quad \\
    $^1$Georgia Institute of Technology\quad
    $^2$
    Facebook AI Research\\
}

\maketitle
\ificcvfinal\thispagestyle{empty}\fi

\begin{abstract}

\objnavfull (\objnav) is an embodied task wherein agents are to navigate to an object
instance in an unseen environment. 
Prior works have shown that end-to-end \objnav agents that use vanilla visual and recurrent modules,
\eg a CNN+RNN, perform poorly due to overfitting and sample
inefficiency. 
This has motivated current state-of-the-art methods to mix analytic and learned components and operate on explicit spatial maps
of the environment. 
We instead re-enable a generic learned agent by adding auxiliary learning tasks and an exploration reward.
Our agents achieve $24.5\%$ success and $8.1\%$ SPL,
a 37\% and 8\% relative improvement over prior state-of-the-art, respectively, on the Habitat ObjectNav Challenge~\cite{habitat_challenge2020}.
From our analysis, we propose that agents will act to simplify their visual inputs so as to smooth their RNN dynamics, and that auxiliary tasks reduce overfitting by minimizing effective RNN dimensionality; \ie a performant \objnav agent that must maintain coherent plans over long horizons does so by learning smooth, low-dimensional recurrent dynamics. 

Site: \href{https://joel99.github.io/objectnav/}{{\tt joel99.github.io/objectnav/}}

\end{abstract}
\vspace{-10pt}

\section{Introduction}
\vspace{-0.06in}

Consider how a robot placed in a novel home environment should find a target object, \eg \myquote{Find a chair}.
This task, known as \objnavfull (\objnav), requires the agent to search through the unseen environment and,
upon seeing a goal, navigate around obstacles to reach it.

Current state-of-the-art \objnav agents~\citep{chaplot2020object,habitat_challenge2020} build explicit spatial maps of the environment and leverage a mix of analytic and learned planning on top of these maps.
This is in contrast to the state-of-the-art methods in \pointnav, a related
task where agents navigate to specified goal coordinates (instead of object
categories). In \pointnav, the best method~\citep{wijmans_iclr20} scales a generic architecture (\ie a CNN and
RNN) to $2.5$ billion
frames of experience. This approach outpaces methods with explicit
spatial grounding, to the point of essentially solving \pointnav in Habitat~\cite{savva_iccv19}.

In this work, we explore how to enable such a generic agent for
\objnav.
\objnav is considerably more challenging than \pointnav; the agent that nearly solves \pointnav only achieves $6\%$ success on \objnav~\citep{habitat_challenge2020}.
While \pointnav agents can use the onboard GPS+Compass sensor as a compact
representation to measure progress towards the goal, \objnav agents cannot; they instead need to be competent at exploration since the target object can be anywhere
in the environment.
As such, a vanilla CNN+RNN agent tasked with learning these complex
representations for \objnav is likely to be under-equipped with only an RL
reward, even if the reward is shaped to encourage navigation towards a
goal~\cite{habitat_challenge2020}.
Even with an additional exploration reward and no other feedback to 
learn better environment representations, efficient exploration is hard~\citep{chaplot_iclr20}.

\begin{figure}[t]
    \centering
    \includegraphics[width=0.46\textwidth]{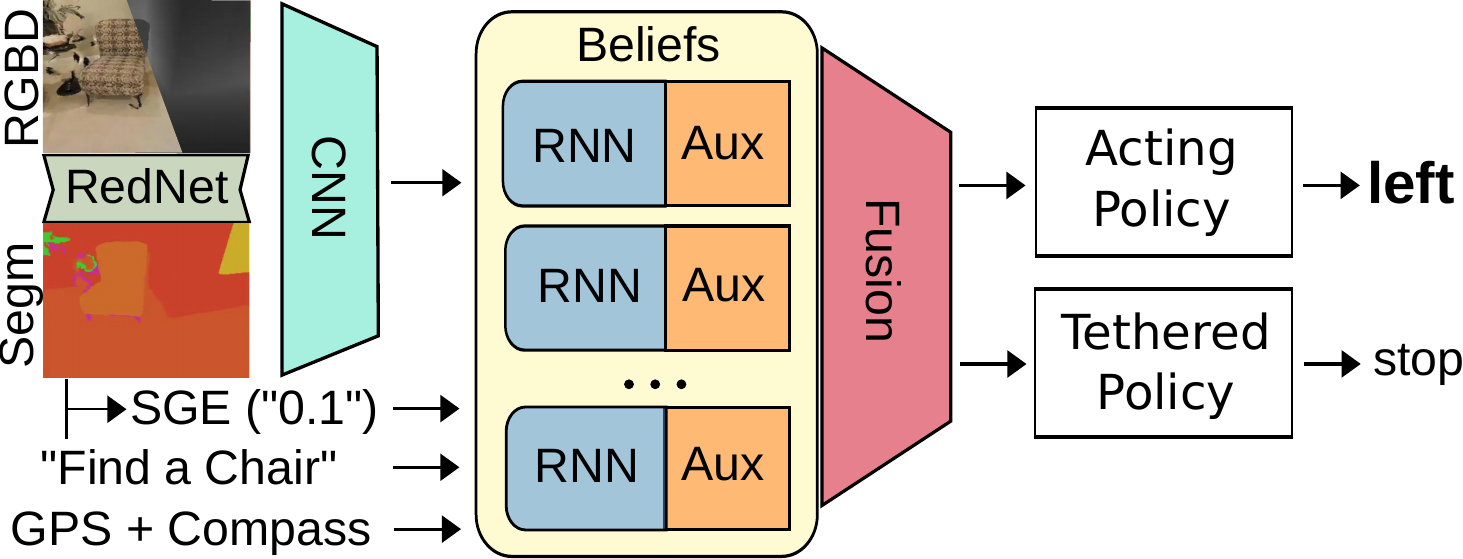} \\
    \vspace{5pt}
    \caption{Overview of the \objnav agent architecture. The agent receives RGBD input, a GPS+Compass sensor, and must navigate to a goal instance. Building on~\citep{ye2020aux},
    we introduce new auxiliary tasks, a semantic segmentation visual input,
    a Semantic Goal Exists (SGE) feature that describes the fraction of the frame occupied by the goal object,
    and a method for tethering secondary policies that learns from its own reward signal with off-policy updates.
    We encourage the acting policy to explore, and the tethered policy to perform efficient \objnav.}
    \vspace{-15pt}
    \label{fig:arch}
\end{figure}

Our approach builds on a recent advance in \pointnav~\citep{ye2020aux}
that combines simple architectures with auxiliary learning tasks to greatly
improve sample complexity in visually complex environments.
Specifically, 1) we update the agent's inputs to incorporate egocentric
semantic segmentation and a feature that explicitly describes how much
of the target object class is in view.
2) We additionally introduce three new auxiliary learning tasks: two general task for
learning an inverse dynamics model and one exploration-centric task that predicts map coverage,
and
3) to teach the agent to explore, we add an exploration term to the reward.

However, exploration is different than \objnav and we find that agents trained with an exploration reward will continue to explore after locating goal instances. To mitigate this task mis-alignment, we propose a `tethered-policy' multitask learning technique. In this method, we learn a second policy adapted only for the sparse \objnav reward while still acting via a primary policy given the exploration reward. Tethering allows faster adaptation to the sparse \objnav reward, which in turns improves agent efficiency over simple fine-tuning of a single policy on the sparse reward.

Our results show that end-to-end learning can
achieve state-of-the-art results in \objnav once equipped with
generic representation learning and exploration objectives. 
Nonetheless we are far from human-like competence at the task. To guide future work, we analyze various aspects of our agent, but center in particular around its \emph{recurrent dynamics}. We propose that zero-shot transfer to RedNet segmentation is greatly degraded by the chaotic dynamics it causes in agent RNNs, that several failure modes are caused by lapses in agent memory, and that auxiliary tasks reduce overfitting by constraining the effective dimensionality of agent RNNs.

Concretely, our contributions are:
\begin{compactenum}[1)]
    \item Objectives to train a simple CNN+RNN architecture that lead to
        24.5\% success on \objnav (+37\% relative improvement over prior state-of-the-art). 
        This suggests that, despite their current prevalence,
        explicit maps
        need not be necessary to
        learn complex Embodied AI tasks.
    \item A method for tethering secondary policies to an agent. Tethered policies learn from separate reward signals with off-policy updates. We use a tethered policy with no exploration reward to achieve 8.1\% SPL (+8\% relative improvement over prior work).
    \item An analysis of the agent. Through examination of the agent's behavior, representations, and recurrent dynamics, we find a) the agent seeks out simple visual inputs corresponding to smoother RNN dynamics, b) agent failures are often related to agent memory, and c) auxiliary tasks regularize agent RNNs by constraining their effective dimensionality. 
    We hypothesize a key component of performant \objnav agents will be the ability to plan with smooth, low-dimensional recurrent dynamics.

\end{compactenum}

\vspace{-0.06in}
\section{Approach}
\vspace{-0.06in}

\xhdr{\objnav Definition and Dataset}.~\citep{anderson_arxiv18}  In \objnav,
an agent must navigate to an instance of a specified object category in an
unseen environment. The agent does not receive a map of the environment and must
navigate using its (noiseless) onboard sensors: an RGBD camera, a GPS+Compass
sensor which provides location and orientation relative to start of episode.
The agent also receives a goal object category ID. The full action space is
discrete and consists of \textsc{move\_forward}, \textsc{turn\_left}, \textsc{turn\_right},
\textsc{look\_up}, \textsc{look\_down}, and \textsc{stop}. We experiment with
both the full 6-action setting and a restricted 4-action setting that excludes
\textsc{look\_up} and \textsc{look\_down}. To our knowledge, no other works have
considered the 6-action setting (perhaps due to the increased complexity of
specifying tilt actions in $2$D map-based planners). The agent must stop at a
location within 1$m$ of an object of the specified category and be able to view
the object from that location.  There is a $500$ step limit to succeed on the
episode.

We experiment on the Matterport dataset (MP3D~\citep{mp3d}), which has $90$
scenes and $40$
labeled semantic object categories. There are $21$ goal categories, as specified
in the Habitat 2020 Challenge~\citep{habitat_challenge2020}. 
We train our agent with a $3$D GPS (\ie with vertical localization) to encourage
the agent to reason about 3D exploration, as a considerable number
of MP3D scenes incorporate elevation changes. However, we evaluate the agent
with a $2$D GPS for compatibility (by zero-ing vertical axis) with the Habitat Challenge parameters.
We provide a comparison of these settings in~(\secref{sec:2d_3d}).

\xhdr{Additional Features}. We accelerate agent learning by providing %
two semantic features: semantic segmentation (segm.) for the visual input, and
a ``Semantic Goal Exists'' (SGE) scalar which equals the fraction of the visual
input that is occupied by the goal category (computed from the semantic segm.).
During training, we use the ground truth semantic segm.~directly from the MP3D
annotations. During evaluation, the segm. is predicted from RGBD using a
RedNet~\citep{jiang2018rednet} model finetuned to predict the $21$ goal categories. The RedNet model 
is taken off-the-shelf (trained on SUNRGBD \citep{song2015sun}) and
finetuned on $100$k randomly sampled forward-facing views from MP3D. We provide
additional details in~\secref{sec:rednet}.
The SGE feature distills the domain knowledge that the agent should navigate to
the goal once it is seen; this knowledge is built into the
planners in prior works~\citep{chaplot2020object,liang2020sscnav}.

\vspace{-0.06in}
\subsection{Agent Architecture}
\vspace{-0.06in}

Our agent uses the split-belief architecture introduced in~\citep{ye2020aux},
shown in~\figref{fig:arch}. This approach first embeds all sensory inputs
with feed-forward modules. The visual inputs are first downsampled by $0.5$ \ie from $480 \times 640$ to $240 \times 320$, and the semantic segm.~channel is also projected to 4D by associating each semantic ID with a learnable vector. After this preprocessing, visual inputs are fed through
a ResNet18~\citep{he_cvpr16}. 
The goal object ID is directly
embedded into a $32$D vector. The embedded visual and goal vectors are
concatenated with the GPS-Compass and SGE inputs to form an observation embedding. Next, a set of recurrent
``belief modules'' integrate these observation embeddings over time.
These belief modules are independent GRUs~\citep{cho_emnlp14}, each
associated with a separate auxiliary task; Ye \etal~\citep{ye2020aux} proposes that the
split modules, as opposed to one larger recurrent network, enable orthogonal
auxiliary tasks to be learned while minimizing task conflict. We term output cell
states from all GRUs as ``beliefs.''
Beliefs are fused using an attention layer conditioned on the observation embedding.
The fused belief is then directly used in a linear actor-critic policy head. We refer to this agent as the base agent.

\vspace{-0.06in}
\subsection{Learning Signals}
\vspace{-0.06in}
\xhdr{Rewards}.
The base agent receives a sparse success reward $r_{\text{success}}$, a slack
reward $r_{\text{slack}}$ to encourage faster goal-seeking, and an exploration
reward $r_{\text{explore}}$. Measuring exploration is a dense indicator of progress,
and encourages an intuitively prerequisite skill for \objnav. We use a
visitation-based coverage reward, in which we first divide the map into a voxel
grid with $2.5m \times 2.5m \times 2.5m$ voxels and reward the agent for
visiting each voxel.
This form of visitation-based exploration bonus was found to work well for
map-based agents~\citep{ramakrishnan2020exploration}. We smooth $r_\text{explore}$ by decaying it by the number of steps the agent has spent in the voxel (visit
count $v$). Further, to ensure the agent eventually prioritizes \objnav, we decay $r_\text{explore}$ based on episode timestep $t$ with a decay constant $d = 0.995$.
In summary:
\begin{subequations}
\begin{align}
    r_{\text{total}} &= r_{\text{success}} + r_{\text{slack}} + r_{\text{explore}}\\
    r_{\text{success}} &= 2.5 \quad\quad\,\,\, \text{on success} \\
    r_{\text{slack}} &= -10^{-4} \quad \text{per step} \\
    r_{\text{explore}} &= 0.25 \times \frac{d^{t}}{v}
\end{align}
\end{subequations}

\begin{table*}[t]
    \centering
        \begin{tabular}{@{}lccccc@{}}
            & \multicolumn{2}{c}{\textsc{val}} & & \multicolumn{2}{c}{\textsc{test-std}} \\
            \cline{2-3}
            \cline{5-6}
            & Success \% $(\mathbf{\uparrow})$ & SPL \% $(\mathbf{\uparrow})$ & &
            Success \% $(\mathbf{\uparrow})$ & SPL \% $(\mathbf{\uparrow})$ \\[0.05in]
            \toprule
        \rownumber 4-Action &
        \boldsymbol{
            $34.4 $
         }\scriptsize{\boldsymbol
             {$\pm 2.0$}
         }
         & 
         $9.58 $\scriptsize{$\pm 0.75$}
        & &
         19.9 & 6.7
        \\
        \rownumber 4-Action + RedNet FT &
        \boldsymbol{
            $33.1 $
         }\scriptsize{\boldsymbol
             {$\pm 2.0$}
         }
        & 
        $6.89 $\scriptsize{$\pm 0.56$}
        & &
        - & -
        \\
        \rownumber 6-Action &
         $30.8 $\scriptsize{$\pm 1.9$}
         & 
         $7.60 $\scriptsize{$\pm 0.64$}
        & &
        - & -

        \\
        \rownumber 6-Action + Rednet FT &
         \boldsymbol{
            $34.6 $
         }\scriptsize{\boldsymbol
             {$\pm 2.0$}
         }
         & 
         $7.93 $\scriptsize{$\pm 0.64$}
        & &
         $\mathbf{24.5}$ & 6.4
        \\

        \rownumber 6-Action + Tether &
        $26.6 $\scriptsize{$\pm 1.9$}
        & 
        $9.79 $\scriptsize{$\pm 0.82$}
        
        & &
        - & -

        \\
        \rownumber 6-Action + Tether + RedNet FT &
        $30.3 $\scriptsize{$\pm 1.9$}
        & 
        \boldsymbol{
            $10.8 $
         }\scriptsize{\boldsymbol
             {$\pm 0.84$}
         }
        
        & &
         21.1 & $\mathbf{8.1}$
        \\[0.01in]
        \cdashline{1-6}

        \rownumber E2E Baseline (DD-PPO~\citep{wijmans_iclr20}) & - & -
        & &
         6.2 &  2.1
        \\

        \rownumber SemExp~\citep{chaplot2020object} &
         - & -
        & &
         17.9 & 7.1
        \\

        \rownumber SRCB-robot-sudoer &
         - & -
        & &
        14.4 & 7.5
        \\
        \bottomrule
        \end{tabular}
    \vspace{5pt}
    \caption{Primary variants on the MP3D \textsc{val} and \textsc{test-std} splits, along with prior SoTA from the Habitat Challenge
    leaderboard~\citep{habitat_challenge2020}. $\pm$ intervals provide 95\% CI; \textsc{Val} bold values are significantly better than non-bold values ($p < 0.05$, paired t-test), \textsc{Test} bolding is for emphasis. Our auxiliary task and exploration enabled agent (row 1) outperforms prior state of the art success (over row 7) and triples the vanilla learning agent (row 6). Fine-tuning the agent with RedNet segmentation reverses the ranks of 4 and 6-action agents (rows 1 and 3 vs 2 and 4, analysis in~\secref{sec:gt_results}), helping the 6-action agent to achieve a new state-of-the-art 24.5\% success. Finally, tuning on the sparse \objnav success reward helps the agent set state-of-the-art 8.1\% SPL (row 6 vs 9).
    }
    \vspace{-15pt}
    \label{tab:primary}
\end{table*}

\xhdr{Tethering to an Exploration Policy}. Using exploration to guide
behavior can distract the agent from properly terminating at the goal point,
\eg when it is standing next to the goal but can instead choose to continue
to explore a wide open space (we observe this in~\secref{sec:error_anlaysis}). To alleviate this pathology, we introduce a
tethered-policy method which attaches an additional policy head to the fused belief
output.
Each of the two policy heads can be fed their own desired rewards, and
the agent can act according to any mix of the two policies. Note this means that each policy share the same action space. We implement a simple
schedule in which the agent first acts with a policy optimized with the full
shaped reward ($r_\text{total}$) for an initial period of training,
and then with a second policy which only uses the sparse $r_\text{success}$. The first phase is intended to teach the agent how to explore, and the second to encourage efficient \objnav. We use
importance-weighted VTrace~\citep{espeholt2018impala} returns to update the
non-behavioral policy to account for experience being off-policy from its
perspective (details in~\secref{sec:tethered}). Tethering is similarly motivated as SAC-X~\citep{riedmiller2018sacx}, but implemented as a lightweight extension to a standard RL agent.

\xhdr{Auxiliary Tasks}. We use  6 auxiliary tasks. As in Ye \etal~\citep{ye2020aux}, we use 2 instances of \cpc~\citep{guo_arxiv18}, a self-supervised contrastive task, at horizons of $k=\{4,16\}$ steps. We also use PBL~\citep{guo2020pbl} with a horizon of $k=8$ steps. These tasks use agent states to make predictions about the environment.

We also introduce two general-purpose inverse tasks, which predict agent actions given environment transitions: Action Distribution Prediction (ADP) and Generalized Inverse Dynamics (GID). ADP and GID both predict actions taken between two observations $k$ frames apart (\ie $\{a_{[t:t+k]}\}$), and are both conditioned on the belief at the first frame $h_t$ and the visual embedding (\ie CNN output) $\phi_{t+k}$. ADP uses a 2-layer MLP to directly predict an action distribution and evaluates the KL-divergence between this prediction and the empirical distribution of the next $k$ actions:
\begin{align}
    L_{\text{ADP}} &= KL(\text{MLP}(h_t, \phi_{t+k}), a_{[t:t+k]})
\end{align}
GID predicts individual actions using a separate GRU with state $g^{\text{GID}}$ and two linear layers $f, f'$. The GRU is initialized with the same inputs as in ADP:
\begin{align}
    g_{t}^{\text{GID}} &= f(h_t, \phi_{t+k})
\end{align}
The GRU is updated with actions taken:
\begin{align}
    g_{t+i}^{\text{GID}} &= \text{GRU}(a_{t+i-1}, g_{t+i-1}^{\text{GID}}) \\
    L_{\text{GID}} &= \sum_{i=1}^{k-1}\text{CrossEnt}(f'(g_{t+i}^{\text{GID}}), a_{t+i})
\end{align}

Finally, we introduce an exploration-specific Coverage Prediction (CP) task that leverages the GPS sensor provided to the agent. CP follows a similar structure as GID, \ie using a GRU for sequence prediction. However, this GRU's initial state is conditioned not on a visual embedding, but on the number of steps the agent has spent in its current voxel, $v(s_t)$. This conditioning helps the GRU anticipate how close the agent is to the edge of the voxel. The GRU predicts the change in coverage at each of the next $k$ steps:
\begin{align}
    g_{t}^{\text{CP}} &= f''(h_t, v(s_t)) \\
    g_{t+i}^{\text{CP}} &= \text{GRU}(a_{t+i-1}, g_{t+i-1}^{\text{CP}}) \\
    L_{\text{CP}} &= \sum_{i=1}^k \text{CrossEnt}(f'''(g_{t+i}^{\text{CP}}), \Delta_\text{cov}(t, t+i))
\end{align}
where $f'', f'''$ are linear layers, $g^{\text{CP}}$ denote the GRU's state, and $\Delta_\text{cov}(t, t+i)$ denotes the change in coverage, \ie number of new voxels visited between time $t$ and $t+i$\footnote{Note that $\Delta_\text{cov}(t, t+i) \in \{0, \ldots, i\}$}. Perfect performance would require the belief to remember all prior locations. The horizons for ADP, GID, and CP are $k=6,4,16$, respectively; details in~\secref{sec:training_details}.

\section{Results}
\vspace{-0.08in}

In our experiments, we train each of our agents for 8 GPU-weeks (192 GPU-hours), amounting to
${\sim}125$M frames per agent. For the tethered variant, we use the sparse policy for the final 25M frames. The two primary metrics we report are Success and Success weighted by Path Length (SPL), a measure of trajectory efficiency, defined over $N$ episodes as
\[
    \text{SPL} = \frac{1}{N}\sum_{i=1}^N S_i \frac{l_i}{\text{max}(p_i, l_i)}
\]
where $S_i$ is a binary indicator of success on episode $i$, $l_i$ is length of
optimal path, and $p_i$ is length of agent path. We evaluate checkpoints every
${\sim}4$M frames and report metrics from the checkpoint with the highest SPL on~\textsc{val}.

\xhdr{Auxiliary Tasks and Exploration produce an effective \objnav agent}.
We compare against results from the Habitat Challenge 2020 leaderboard in~\tableref{tab:primary}.
We first note the large drop from \textsc{val} to \textsc{test-std} (\eg -14\% success in row 1).
Since we cannot access the test split, we conjecture that the splits were randomly
sampled such that the 9 test scenes were disproportionately challenging.\footnote{ Distribution shift is expected given small \textsc{val, test} splits.} While other works do not highlight this shift, we note that even agents with strong priors
(rows 6, 7) have high performance variance, \eg 5\%
success gaps between the two test splits (\textsc{test-std}, \textsc{test-challenge} \citep{habitat_challenge2020}).
We provide additional commentary and check for agent biases in~\secref{sec:agent_stats}.

Using RedNet segm.~directly (\ie zero-shot transfer from GT~, row 1),
the 4-action agent reaches 19.9\% success (+11\% relative over prior best, row 1 vs 8). 
This is 3x the performance of the E2E baseline that is rewarded for approaching a goal, a method which ``solved'' \pointnav (row 7).

We tune the 4-action agent with RedNet segm.~for an additional 10M frames to account
for the ground-truth-to-RedNet distribution shift, but surprisingly performance
\emph{drops} (row 1 vs 2). Conversely, the 6-action agent which initially performs worse
than the 4-action agent (row 1 vs 3) overtakes the 4-action agent
after finetuning with RedNet (row 2 vs 4). We analyze this reversal in~\secref{sec:gt_results}. Though these tuned agents match in \textsc{val} (row 1 vs 4), the 6-action agent improves on \textsc{test-std}, to 24.5\%,
+37\% relative over prior state-of-the-art (row 4 vs 8).

However, both 4 and 6-Action agents are less efficient than prior methods (SPL column). Qualitative examination of these agents indicate they prefer to wander around goals for some time before stopping at them correctly, likely due to the exploration reward. After finetuning on the sparse \objnav reward through tethered training, SPL rises and Success falls (row 3 vs 5). We show the Success drop is likely due to reduced exploration in~\secref{sec:sparse_exp}. Nonetheless, after tuning this agent on RedNet, we achieve +0.6\% (+8\% relative) over state of the art SPL (row 6 vs 9).

\begin{table}[t]
    \centering
    \resizebox{\linewidth}{!}{
        \begin{tabular}{@{}lccc@{}}
            & Success \% $(\mathbf{\uparrow})$ & SPL \% $(\mathbf{\uparrow})$ \\[0.05in]
            \toprule
\rownumber Base (4-Action) &
\boldsymbol{
    $34.4 $
 }\scriptsize{\boldsymbol
     {$\pm 2.0$}
 }
& 
\boldsymbol{
            $9.58 $
         }\scriptsize{\boldsymbol
             {$\pm 0.75$}
         }
\\
\rownumber \: - Action Distribution Prediction &
\boldsymbol{
    $34.1 $
 }\scriptsize{\boldsymbol
     {$\pm 2.0$}
 }
 & 
 \boldsymbol{
    $9.27 $
 }\scriptsize{\boldsymbol
     {$\pm 0.69$}
 }
\\
\rownumber \: - Generalized Inverse Dynamics &
\boldsymbol{
    $33.7 $
 }\scriptsize{\boldsymbol
     {$\pm 2.0$}
 }
& 
 \boldsymbol{
    $9.05 $
 }\scriptsize{\boldsymbol
     {$\pm 0.69$}
 }
\\
\rownumber \: - Coverage Prediction &
 $32.4 $\scriptsize{$\pm 2.0$} & 
 $8.7s7 $\scriptsize{$\pm 0.69$}
 \\
 \rownumber \: - Semantic Goal Exists &
 $20.3 $\scriptsize{$\pm 1.7$}
 & $4.14 $\scriptsize{$\pm 0.47$}
\\[0.01in]
\cdashline{1-4}
\rownumber  6-Action + Tether &
\boldsymbol{
    $26.6 $
 }\scriptsize{\boldsymbol
     {$\pm 1.9$}
 }
& 
\boldsymbol{
    $9.79 $
 }\scriptsize{\boldsymbol
     {$\pm 0.82$}
 }
\\
\rownumber 6-Action + Sparse Tuning &
 $23.1 $\scriptsize{$\pm 1.8$}
 & $8.43 $\scriptsize{$\pm 0.76$}

\\
            \bottomrule

        \end{tabular}
    }
    \vspace{5pt}
    \caption{Ablations on \textsc{val}. Bold values are significantly better than unbolded values in the same group. New auxiliary tasks provide moderate performance gains, and tethered training preserves Success better than directly tuning on sparse \objnav reward.}
    \label{tab:ablation}
\end{table}

We provide ablations for our design choices in~\tableref{tab:ablation}. Our auxiliary task ablations are compared against the 4-action agent (rows 1-4). These tasks provide modest gains in both metrics (<1\% for ADP, 1\% GID, 2\% CP). Next, ablating the SGE sensor drops performance by a large amount (-14\%, row 1 vs 5). In~\secref{sec:aux_sge}, we note that SGE is much more effectively incorporated as a feature than as an auxiliary task.
Finally, we also compare tethered policy training against direct finetuning (row 6 vs 7) for the 6-action agent.
While direct finetuning also improves SPL (vs SPL reported in row 3 in~\tableref{tab:primary}), tethered training is overall more effective at preserving agent performance.

\begin{table}[t]
    \centering
    \resizebox{0.975\columnwidth}{!}{
        \begin{tabular}{@{}lccccc@{}}
            & \multicolumn{2}{c}{Success \% $(\mathbf{\uparrow})$} & & \multicolumn{2}{c}{SPL \% $(\mathbf{\uparrow})$ } \\
            \cline{2-3}
            \cline{5-6}
            &
            \textsc{train} & \textsc{val}
            & &
            \textsc{train} & \textsc{val} \\[0.03in]
            \toprule
        \rownumber 4-Act &
        50.3 (36.0) & 
        43.3 ($\mathbf{34.4}$) %
        & &
        18.1 (12.4) & 
        12.3 (9.6) %
        \\
        \rownumber 6-Act &
        56.0 (21.7) & $\mathbf{58.0}$ (30.8) %
        & &
        21.5 (8.2) & 16.9 (7.6) %
        \\
        \rownumber 6-Act + Tether &
        54.0 (27.3) & 52.2 
        (26.0) %
        & &
        27.9 (11.5) & $\mathbf{26.0}$
        ($\mathbf{9.8}$) %
        \\
    \bottomrule
    \end{tabular}}
    \vspace{5pt}
    \caption{Performance on a 300-episode subset of \textsc{train} and \textsc{val}
        splits, reported as ``with GT segmentation (with RedNet segmentation)''. Best metrics are bolded for emphasis.
        In both splits, all agents
        degrade significantly with RedNet segmentations, but 6-action agents
        more so. Under GT segmentation (numbers outside parentheses), agents are
        minimally overfit, though SPL does degrade slightly from \textsc{train}
        to \textsc{val}. In the GT setting, 6-action agents outperform 4-action
        agents.}
    \vspace{-15pt}
    \label{tab:gt_segm}
\end{table}
\vspace{-0.08in}
\subsection{Stable segmentation is critical to untuned 6-action agents.}
\vspace{-0.08in}
\label{sec:gt_results}
We found that 4-action agents suffer a performance drop after finetuning on RedNet, while 6-action agents improve.
To understand why, we first observed that the RedNet is greatly overfit (\secref{sec:rednet}), we decouple our agent from RedNet by evaluating agent performance on both \textsc{train} and \textsc{val} with GT and RedNet segm.~, shown in~\tableref{tab:gt_segm}. We first note that most agents are minimally overfit to \textsc{train}, with a moderate impact to SPL and $<3{\%}$ drop in success. RedNet Seg. degrades performance significantly, but in particular, RedNet Seg. hurts 6-action agents more, even on \textsc{train}, where the effects of an overfit RedNet are reduced. This suggests 6-action agents are more sensitive to RedNet statistics in particular, and hence able to overtake the 4-action agent once finetuned with RedNet.

We examine agent trajectories to understand this sensitivity (videos provided in supplement). We first find
that during training with GT semantics, 6-action agents learn to navigate
while spending most steps facing downwards towards the floor (while the 4-action
agent can only look straight ahead). This happens despite there being minimal
obstacles on the floor while looking straightforward
offers greater chance to see more distant goals.
Intuitively, the visual input provided by the floor is simpler than the
 view of a room. We thus posit a stability hypothesis: the 6-action
agent learns to exploit a \emph{simpler signal to enable stabler recurrent dynamics}. That is, because it is harder to maintain memory over
long term trajectories when faced with noisy inputs, agents will simplify
their input even at slight cost to information gained. Such a hypothesis may relate to results on improved generalization when restricting the visual
field of a gridworld agent~\citep{Hill2020Environmental} and when withholding
inputs from parts of a recurrent module~\citep{goyal2020recurrent}. This hypothesis
suggests that standard RNN gating is imperfect at preventing
input-induced information decay in the RNN state, and presents an avenue for future work.

We next observe that RedNet segm.~is low quality and unstable, flickering across consecutive frames and without well-defined object boundaries. These are expected failures: poor segm.~is due in part to the noisiness of the underlying MP3D meshes, and consistent segm.~is a broader challenge~\cite{varghese2020unsupervised}.
Predicted segm.~of frames of the floor are particularly poor, likely since RedNet is trained with front-facing views. Consequently, the 6-action agent behavior was very erratic, with repeated alternating turning actions. If the 6-action agent did learn to exploit the floor for simple visual inputs, it would have a reduced incentive to filter noisy frontal views, and thus would be particularly vulnerable to degraded segm. Then, one possible reason to explain earlier trends is that 1. RedNet overfitting implies that previously general agents which are tuned with RedNet will in turn overfit, 2. Untuned 6-action agents avoid this overfit transfer but suffer significantly more from unstable semantics, and 3. Tuned 6-action agents overfit, but gain performance by maintaining stable behavior while receiving unstable semantics. We quantify this instability in~\secref{sec:action_ent}. The fact that tuning agents with RedNet improves agent performance suggests robustness to noisy inputs is partially learnable, though the agent will not do so if it has a simpler option (facing the ground).

\vspace{-0.08in}
\section{Agent Analysis}
\vspace{-0.08in}

These agents demonstrate promising performance ($\sim$ 55\% val success) in navigating large, complex environments, especially when granted GT segm.
Nonetheless, 
their performance is far from perfect, and they still display qualitative failure modes that a human would not have, such as repeated circling inside a room. In this section we analyze agent behavioral modes, knowledge, and internal dynamics to inform future directions towards human-level \objnav, and to develop an intuition for how a competent \objnav agent operates. We scope our analysis to GT segm.~and the base 6-action agent, unless otherwise noted. 

\vspace{-0.08in}
\subsection{Behavioral Analysis}
\vspace{-0.08in}
\label{sec:error_anlaysis}

\begin{figure}[t]
    \centering
    \includegraphics[width=0.45\textwidth]{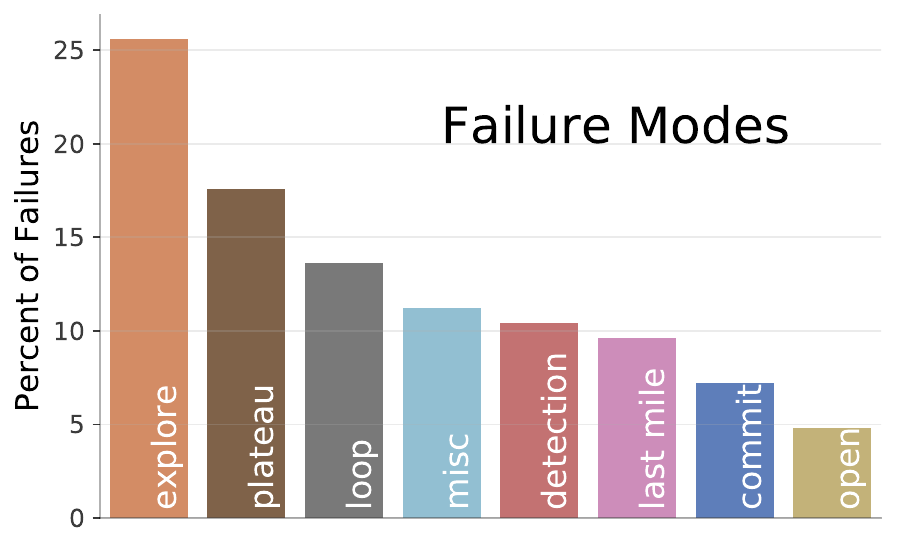}
    \caption{Breakdown of failure modes for base 6-action agent in 300 episodes of \textsc{val}.}
    \label{fig:failure_breakdown}
    \vspace{-20pt}
\end{figure}

In our behavioral analysis we aim to understand: How can \objnav agents approach 1.0 Success? To identify prominent modes of the remaining failures, we conduct a qualitative coding of agent behavior for both the base and tethered 6-action agent.
Specifically, we sample 300 validation episodes and manually label the failures (there are 125 for base, 151 for tether) and group them according to common trends. %
A subset of failure modes are described in~\tableref{table:failure_desc} and their relative prevalence for the base agent is shown in~\figref{fig:failure_breakdown}. 
A full list of failure modes are in \secref{sec:error_full}, and example videos are available in the supplement.

\begin{table}[h]
    \centering
        \begin{tabular}{p{2cm}p{6cm}}
            Name & Description \\[0.05in]
            \toprule
        Explore & 
         A generic failure to find the goal despite steady exploration. Includes semantic failures \eg going outdoors to find a bed.
        \\
        Plateau & 
        Repeated collisions against the same piece of debris causes a plateau in coverage. Includes debris which traps agent in spawn.
        \\
        Loop & 
        Poor exploration due to looping over the same locations or backtracking.
        \\
        Detection &
        Despite positive SGE, the agent does not notice nor successfully navigate to the goal.
        \\
        Last mile &
        Gets stuck near the goal.
        \\
        Commitment &
        Sees and approaches the goal but passes it.
        \\
        Open &
        Explores an open area without any objects.
        \\ 
        \bottomrule
    \end{tabular}
    \caption{Description of prominent failure modes.}
    \label{table:failure_desc}
\end{table}

We find that agent failures are diverse and beyond only unsuccessful exploration, which at best comprises `explore' and `loop'. They get stuck (`plateau' and `last mile'), suffer pathologies that might be attributed to the exploration reward (`commitment' and `open'), and deal with a handful of issues due to scene quality (labeled as `misc' here). However, some failure modes do preclude others.
For \eg, an agent which is trapped by reconstruction debris at spawn (`plateau') is still liable to not find the goal due to general exploration issues. Additionally, the agent does not make panoramic turns to identify promising goal locations, as a human might. We attribute this to the visitation-based exploration reward; a different reward which promotes simply viewing more areas instead of visiting them may mitigate this.
Alternately, non-episodic exploration rewards may promote the appropriate behavior with less misalignment with \objnav.

These exploration-related pathologies are not cured by swapping to the sparse \objnav reward, which degrades success (\tableref{tab:ablation}, row 6 and 7). Tethered agents explore worse, and behavioral coding finds a new dominant failure mode wherein the agent quits early (see~\secref{sec:sparse_exp}).

\vspace{-0.08in}
\subsection{Probing Learned Knowledge}
\label{sec:probes}
\vspace{-0.08in}

\begin{figure}[t]
    \centering
    \includegraphics[width=0.45\textwidth]{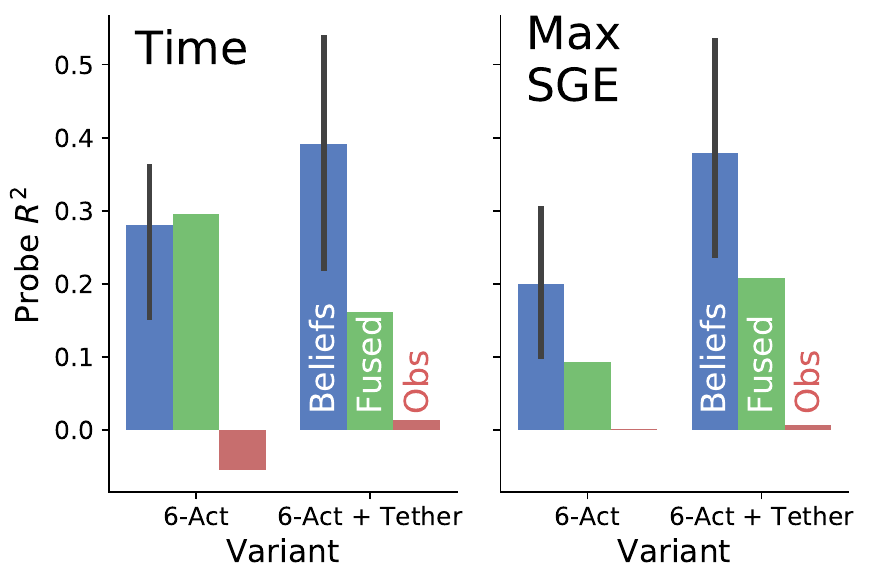} \\
    \caption{We probe time and max SGE seen on each belief, the fused belief, and the observation embedding (``Obs'') as a baseline. Error bars show 95\% CI across single beliefs. Both concepts are encoded, though imprecisely and with high variance across beliefs.}
    \vspace{-10pt}
    \label{fig:probes}
\end{figure}

How well do agent representations match our intuitions? Does its behavior match represented features? \eg, ``commitment'' failures, which generally occur when the agent spots the goal early in the episode, should only be feasible if the agent maintains a time representation and knows it has time to continue gaining coverage rewards. Conversely, does the agent represent prominent variables that a human might? \eg, ``Commitment'' and ``Detection'' failures both imply that agents do not remember SGE features from prior timesteps.  %
To test these questions, we train linear decoders to probe our agent for time and the most goal seen. We first record representations from 300 episodes on both \textsc{train} and \textsc{val}. We train the decoders using \textsc{train} representations and test them on \textsc{val} representations ($\sim 60-70$K steps each).

We plot decoder performance in~\figref{fig:probes}, aggregating performance across individual beliefs. Though the features are represented across beliefs, there is high variance in probe performance across beliefs. Moreover, probe performance on the fused belief is worse than in individual beliefs. These results reveal semantic differences among individual beliefs and the fused belief, suggesting some abstraction is lost in fusion despite it only being a weighted sum.
Separately, the tethered agent represents Max SGE more prominently than the base agent, matching the intuition that the base agent does not retain goals seen in previous steps perfectly.

With the same approach, we find little evidence of features that a human might use in \objnav: time spent in location, room ID, and distance to goal~(\secref{sec:probes_supp}). Supplying these features directly could improve the agent. 

\vspace{-0.08in}
\subsection{Dynamical Analysis}
\vspace{-0.08in}
Though a behavioral analysis provides a sense of how the agent fails, it does not explain why,
and probing techniques fall short when it is unclear what knowledge an agent might encode.
Through our experiments, dynamics have played an important role -- the 6-action agent looking downward,
acting chaotic with predicted segm., and overtaking 4-action after tuning.
Important failure modes appear dynamic in nature: `loop' failures indicate forgetting of visited locations,
`commitment' failures could be attributed to procrastination through the agent's internal clock.
We were thus motivated to characterize the agent's dynamic computations. 

\xhdr{Belief Dynamics correlate with agent observations}.
Before we study RNN dynamics proper, we first validate that RNN dynamics reflect trajectory dynamics, as represented by sequences of observation embeddings. This helps quantify to what extent agent beliefs are grounded in agent observations and actions; a strong connection would support a stability hypothesis (that agents need smooth inputs for smooth RNN dynamics). Alternately, RNN dynamics could be driven by intermediate computation, and thus the RNNs would conservatively incorporate inputs, \ie through gating, so as to best preserve hidden state information.
We measure this with the curvature of the agent belief trajectories and the observation embeddings, by computing the dot product similarity of successive displacements in a sequence of representations (\citep{henaff2019perceptual}, details in~\secref{sec:curvature}). We report the correlation between each belief curvature and observation curvature.

\begin{figure}[t]
    \centering
    \includegraphics[width=0.36\textwidth]{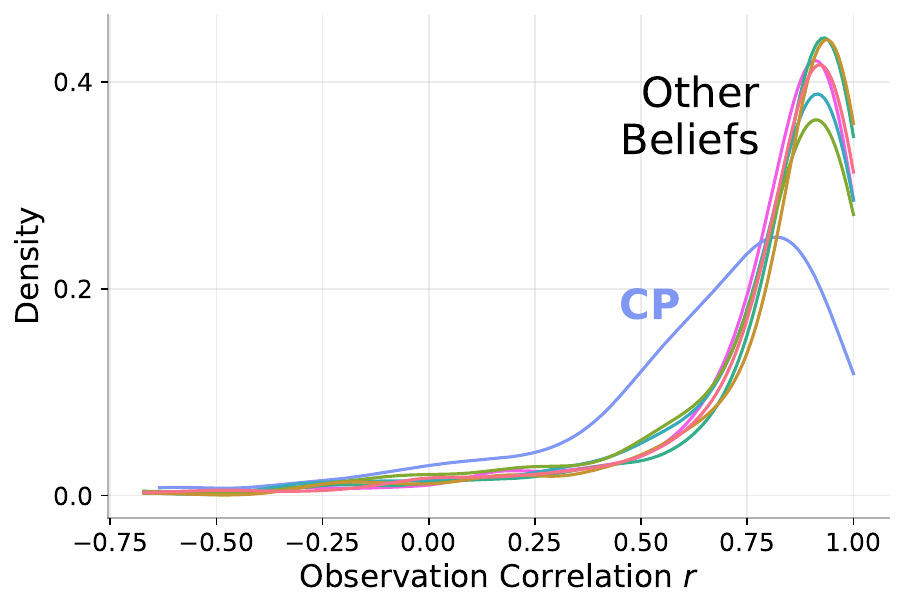} \\
    \caption{
    We measure curvature of observation embeddings and beliefs of the base 6-action agent, plotting the distribution from transitions in 300 episodes. All but CP closely correlate.
    }
    \label{fig:obs_curv}
    \vspace{-6pt}
\end{figure}

\figref{fig:obs_curv} shows that observation and belief curvatures are closely correlated, \ie that varying observations track large changes within agent hidden state. This suggests that RNNs do incorporate most inputs. It is possible non-reactive computation (\eg such as time tracking probed in~\secref{sec:probes}) which occurs independently from observations might not manifest because they are relatively stable; however, the CP belief, which should track coverage variables that are relatively independent from moment-to-moment observations, is less correlated with observations than the other beliefs. This suggests non-reactive computation needn't be hidden.

\begin{figure}[t]
    \centering
    \includegraphics[width=0.36\textwidth]{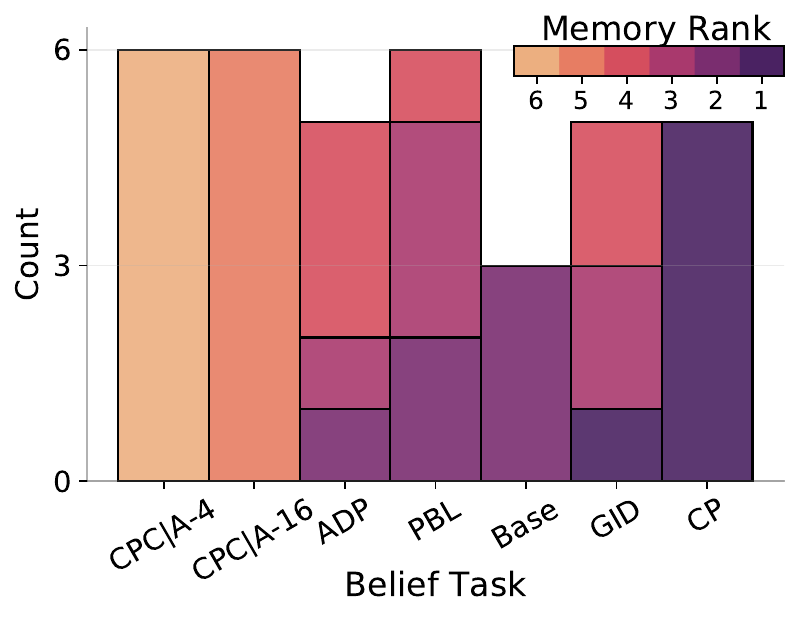}
    \includegraphics[width=0.36\textwidth]{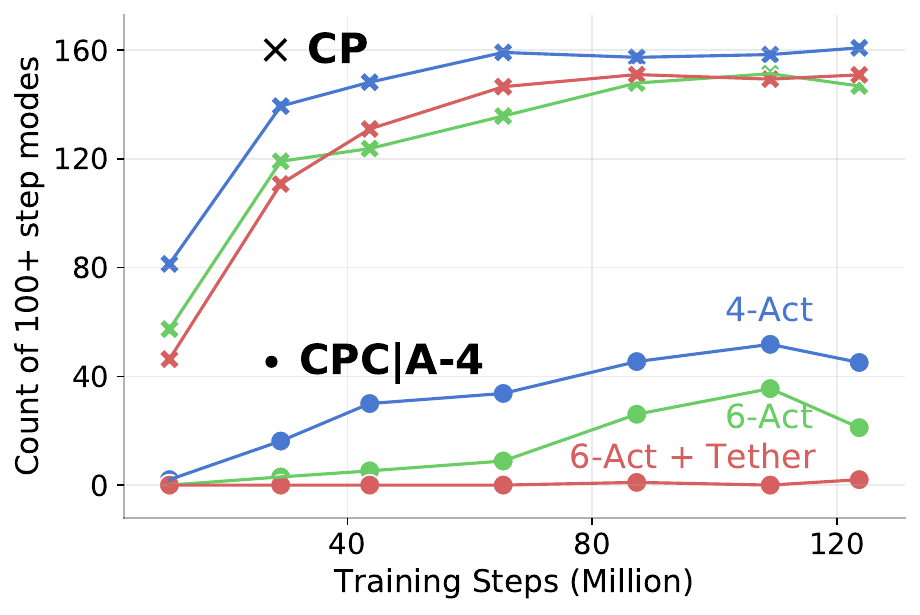}

    \caption{Fixed point (FP) memories. Top: For each of 6 agents (4-Act, 6-Act, 6-Act + Tether, No CP, No ADP, No GID), we sort agent beliefs by the span of their memory. This span is computed by sampling 50 FPs and counting at each FP the number of eigenvalues with time constants above 10 steps. The more eigenvalues past threshold, the larger the span, and the lower the rank.
    Beliefs are labeled by their task, and beliefs without tasks (from ablations) are labeled with \textsc{Base}. We count the number of times a belief appears in a given rank (\eg since \textsc{Base} only appears in 3 agents, it only has 3 counts).
    Consistent rank indicates \emph{relative span is stable.} Bottom: We measure how memory changes over time for the \cpcat4 and CP tasks by counting the number of eigenvalues with time constants above 100 steps at several points during training. 
    }
    \vspace{-15pt}
    \label{fig:memory_rank}
\end{figure}

\xhdr{Auxiliary tasks specify fixed point memory spans}.
We would ideally like to understand what computational structure the agent RNNs use beyond coarse-grained statistics like curvature. To approach this question, we employ fixed point (FP) analysis~\citep{sussillo2013opening,sussillo2015muscle,Golub2018}. FP analysis simplifies the study of a full RNN into a study of its ``slow points,'' where the RNN update is small and well-approximated by a linear function. We provide details on methodology in~\secref{sec:fp_methods}. FPs are typically studied both through their overall layout (the manifold) and their local dynamics. While FP manifolds have been used to directly link RNN computation to dynamical structures, \eg ring attractors, we find our RNNs are much higher-dimensional than those in prior work and thus difficult to clearly classify (notes in~\secref{sec:fp_dim}). This leaves local dynamic structure. At an identified FP $h$, the RNN update is well described by a first order approximation. Let $\Delta h_t := h_t - h$, for $h_{t}$ near $h$, then
\begin{align}
    \Delta h_{t+1} \approx J^{\text{rec}}_h \Delta h_{t}.
\end{align}
Then the decomposition of $J^\text{rec}_h$ assesses FP ``memory.'' Specifically, a delta along an eigenvector $v$ with corresponding eigenvalue $\lambda \ll 1$ would decay to 0, while an eigenvector $v'$ with $\lambda' \approx 1$ would persist indefinitely. A simple method to quantify FP memory is to count the eigenvalues past a threshold value. For interpretability we count eigenvalue time constants instead of the values themselves, as in~\citep{maheswaranathan2020recurrent} (\secref{sec:fp_methods}); we call this quantity a memory span.

We measure and rank these spans for all beliefs of several different agents, producing~\figref{fig:memory_rank}. The auxiliary task corresponding to a belief appears to play a consistent role in specifying FP memory span, and thus in organizing recurrent dynamics, across seeds, action suites (4 vs 6 action), and RL objectives (tether or not). Among the tasks, only CP induces larger spans than the base belief. This matches the intuition that CP could require the tracking of an unbounded number of variables. Separately, the fact that the base belief has larger span than other tasks reveals a potential mechanism by which auxiliary tasks encourage generalization. Note that smaller memories imply there are more dimensions along which RNN state decays towards fixed points, which in turn should imply preference towards lower-dimensional trajectories (even if variable inputs prevent this empirically). Then, the fact that most auxiliary task-augmented beliefs achieve smaller memories than the vanilla ``Base'' belief demonstrates auxiliary tasks may preferentially select for low-dimensional trajectories. Such low-dimensional trajectories could be necessary for generalization; indeed this is theorized to be the reason why monkeys engage low-d dynamics in working memory tasks\citep{cueva2020pnas}. Note that this hypothesis concerns how auxiliary tasks promote generalization in \emph{recurrent} beliefs (\ie beyond general visual representations).

In the bottom figure of~\figref{fig:memory_rank} we plot memory span for several agents at various points in training. Memory span trends upward over training. Consistent with the first plot, \cpcat4 appears to constrain the span to be low. CP, on the other hand, may be constrained by the capacity of the RNN (the belief hidden size is 196). It is possible that an ideal \objnav agent would maintain stable low-d dynamics, but this would need to be reconciled with how such an agent would need to track its path to keep exploring new areas.

\vspace{-0.08in}
\section{Related Work}
\vspace{-0.08in}

Learning an appropriate state representation is a central goal of reinforcement learning. Representations pretrained in static visual tasks, such as the semantic segm.~used in our agent, have been successfully applied to navigation in complex environments~\citep{sax_arxiv18,mousavian2018sem_midlevel}, though results in~\citep{ye2020aux} suggest non-semantic representations may be less effective than end-to-end learned representations. While auxiliary tasks are commonly used to improve agent performance~\citep{mirowski_arxiv16,guo2020pbl}, we believe their role in preventing overfitting to complex environments is underappreciated.

\xhdr{Analyzing Embodied Recurrent Computation}.
Understanding an agent's recurrent computation will be key to understanding their complex behavior. While complex computation resists simple analyses, \eg our probes, the AI community has successfully used sophisticated probing tasks to reveal recurrent representations reflect considerably more environmental knowledge than do static (visual) representations~\citep{weihs2021learning}, and can contain compositional knowledge~\citep{das2020probing}. We have explored a different perspective by analyzing the dynamic rules our agent learns; similar analyses have provided considerable insight in simpler tasks~\citep{aitken2020geometry,sussillo2013opening}. Unfortunately, our agent RNNs are much higher-dimensional than those studied in previous settings and stymies simple fixed point analyses. However, we are encouraged by the different perspective (\eg memory) such dynamic analyses provide, and believe embodied AI provides a good testbed for future development of these dynamic analyses.

\vspace{-0.08in}
\section{Discussion}
\vspace{-0.08in}
Our work uses a recurrent agent with an implicit environment representation to set state-of-the-art for \objnav. 
Our agents are minimally overfit and have yet to saturate training performance, meaning that our approach re-enables scaling as a viable path to improving \objnav. This success relies on auxiliary tasks and an exploration policy, both of which are generic RL ingredients. Future works should consider these ingredients in their generic CNN+RNN baselines, since the default of only providing rewards is demonstrably ineffective in complex environments.
Our analysis indicates that understanding misbehaving agent dynamics is a promising direction, and our results suggest generic learned agents can still drive progress in embodied navigation.

\begin{spacing}{0.8}
\footnotesize
\xhdr{Acknowledgments.}
The authors thank Vincent Cartillier for help preparing the RedNet models. EW is supported in part by an ARCS fellowship. The Georgia Tech effort was supported in part by NSF, AFRL, DARPA, ONR YIPs, ARO PECASE, Amazon.
The views and conclusions contained herein are those of the authors and should not be interpreted as necessarily representing the official policies or endorsements, either expressed or implied, of the U.S. Government, or any sponsor.
\end{spacing}

{\small
\bibliographystyle{ieee_fullname}
\bibliography{strings,main}
}

\clearpage
\newpage
\newpage
\appendix
\renewcommand\thesection{\Alph{section}}
\setcounter{section}{0}
\renewcommand\thefigure{A\arabic{figure}}
\renewcommand\thetable{A\arabic{table}}
\setcounter{figure}{0}
\setcounter{table}{0}
\phantomsection

\section{Appendix}

We use the appendix to elaborate on agent behavior and methodology.

\subsection{Agent Performance against goal distance, category, and scene}
\label{sec:agent_stats}

\begin{figure}[t]
    \centering
    \includegraphics[width=0.4\textwidth]{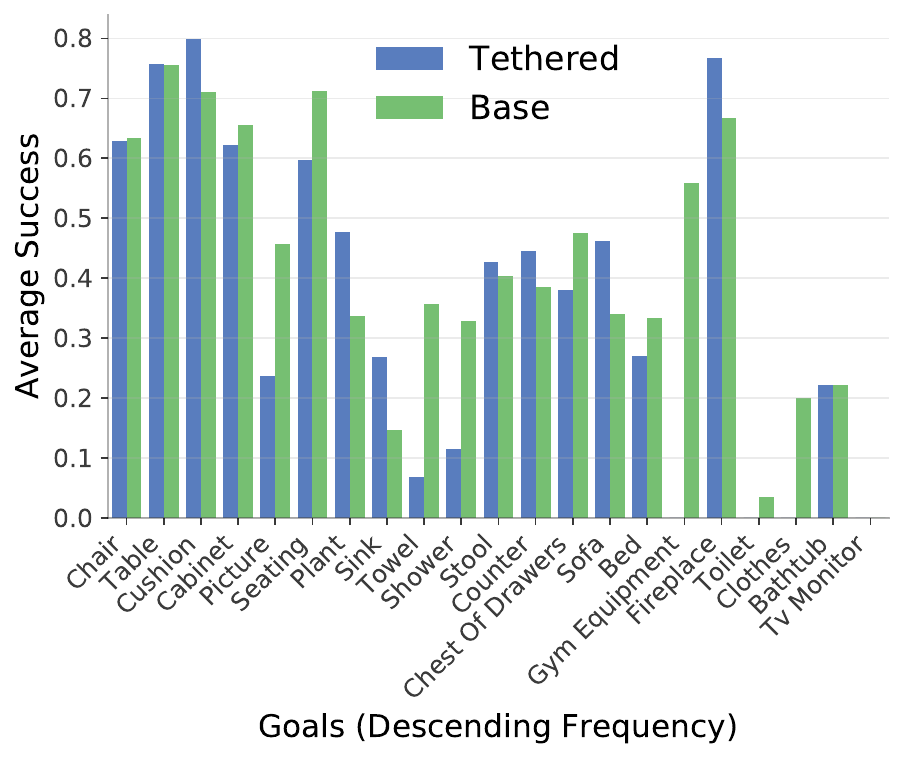} \\
    \caption{While the base agent succeeds on all tasks but ``TV Monitor'', the tethered agent struggles to solve 5 categories.}
    \label{fig:supp_success_goal}
\end{figure}

\begin{figure}[t]
    \centering
    \includegraphics[width=0.4\textwidth]{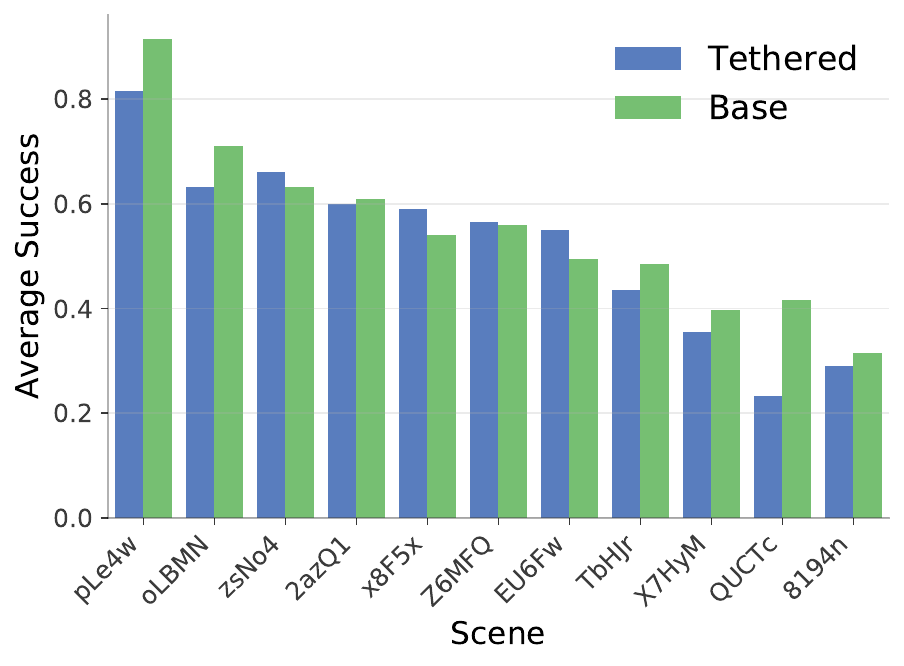} \\
    \caption{While scene difficulty varies, difficulty distribution is not extremely heavy-tailed. \textsc{test-std} shift could result from 1-2 more difficult scenes.}
    \label{fig:supp_success_scene}
\end{figure}

\begin{figure}[t]
    \centering
    \includegraphics[width=0.4\textwidth]{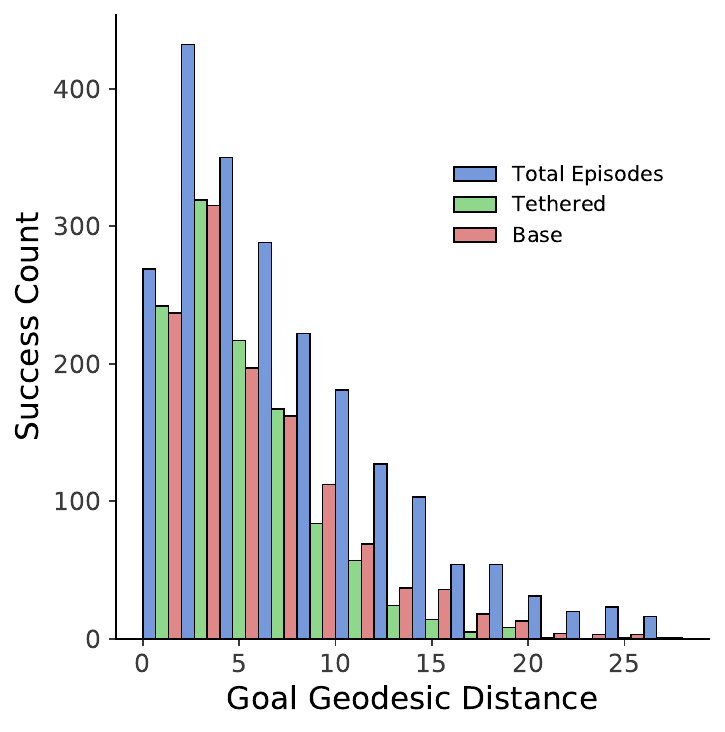} \\
    \caption{Agent success decreases when goals become more distant, and tethered agent is relatively worse at more distant goals.}
    \label{fig:supp_success_d2g}
\end{figure}

Our experiments show large performance variance across \objnav splits. We check for anomalous agent biases by comparing agent success rates conditioned on several episode attributes. Since the large gap between GT and RedNet performance is already established, we provide these plots with GT segmentation.

The base agent has reasonable performance variance across all categories, succeeding less often on rarer categories (\figref{fig:supp_success_goal}) and more distant goals (\figref{fig:supp_success_d2g}). Scene variance (\figref{fig:supp_success_scene}) is wide (0.3 to 0.85), but not particularly heavy-tailed; the \textsc{test-std} drop is likely simply due to increased difficulty. This variance blurs the line that defines state of the art and greatly motivates increasing current \objnav dataset sizes.
 
We also provide the tethered agent's performance; they preferentially succeed on shorter paths relative to the base agent, which can succeed on even very distant goals~\figref{fig:supp_success_d2g}. 

\subsection{Sparse Reward Reduces Exploration and Causes Quitting}
\label{sec:sparse_exp}

\begin{figure}[t]
    \centering
    \includegraphics[width=0.4\textwidth]{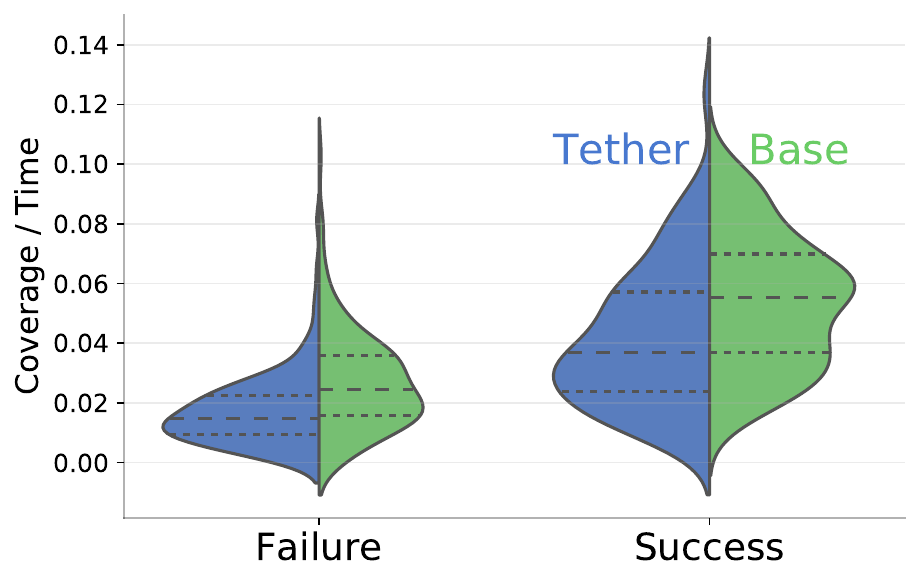}
    \includegraphics[width=0.4\textwidth]{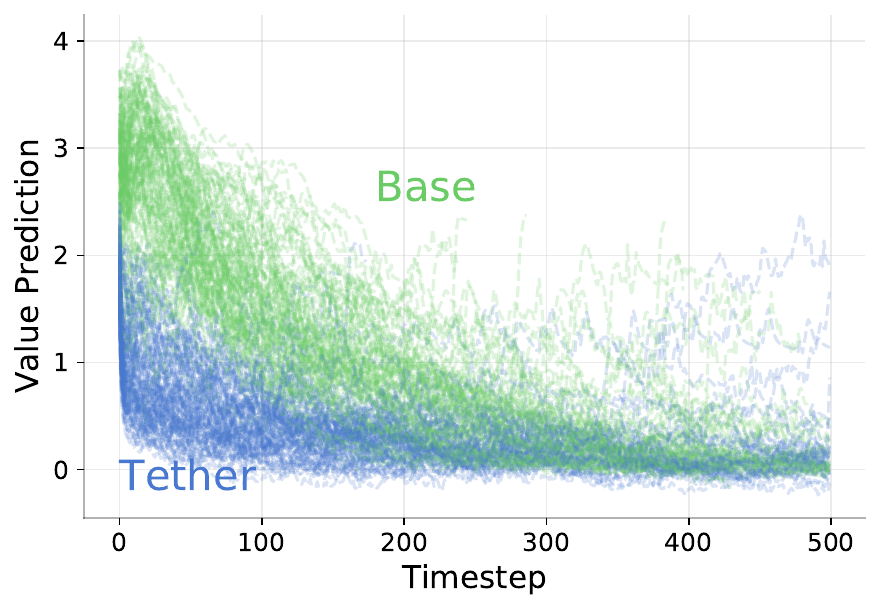}
    \caption{Top: We plot exploration rate (coverage over time) for tether and base agents. In both success and failure conditions, the exploration rate is worse in the tethered agent. Bottom: We plot value function traces for 100 failure episodes. Tether value predictions approaches 0 quickly, but due to estimation error also dips below 0.}
    \label{fig:supp_base_vs_tether}
\end{figure}

The tether agent performs worse than the base agent. We note two primary reasons in~\figref{fig:supp_base_vs_tether}: reduced exploration (top) and early quitting due to estimation error (bottom).
In particular the early quitting is likely a general symptom of using sparse reward. On episodes that are difficult, either due to environmental characteristics (\eg it is outside) or due to goals (\eg clothes, which is difficult at train time), value estimates will occasionally dip below 0 and cause the agent to randomly stop. This may be mitigated by an increased discount factor.

\subsection{Train-Val Gap}
\begin{figure}[t]
    \centering
    \includegraphics[width=0.4\textwidth]{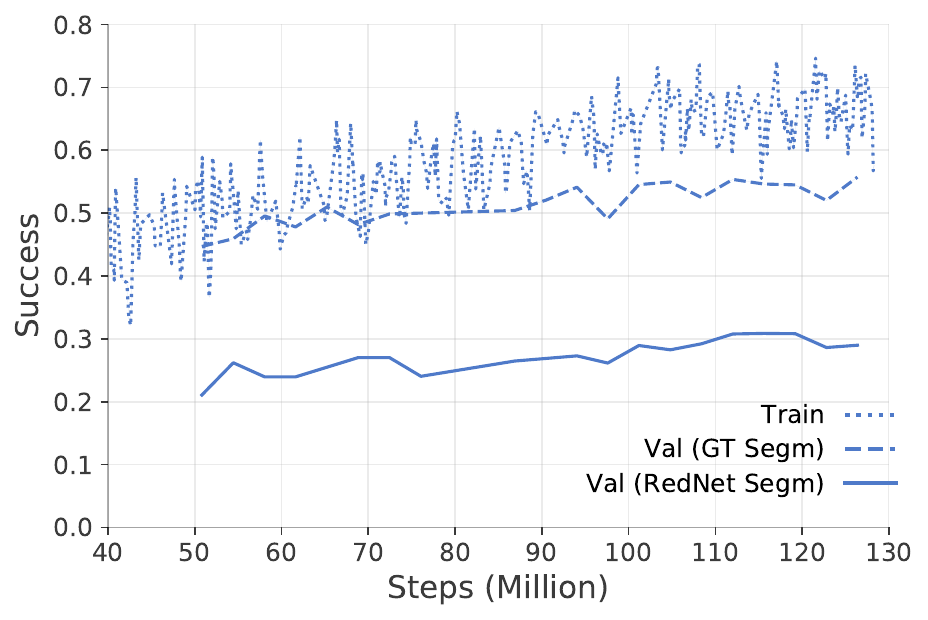}
    \caption{Base 6 Action Agent, training and validation curves. Note: Training curve is taken directly from training logs, \ie it evaluates statistics on the rollout and not on the whole training set.}
    \label{fig:tv_overfit}
\end{figure}

Though~\tableref{tab:gt_segm} suggests virtually no overfitting, our agent does appear to begin moderately overfitting past $0.5$ success in~\figref{fig:tv_overfit}.

\subsection{2D vs 3D GPS Comparison}
\label{sec:2d_3d}

\begin{table}[t]
    \centering
    \resizebox{0.9\linewidth}{!}{
        \begin{tabular}{@{}lccc@{}}
            & Success \% $(\mathbf{\uparrow})$ & SPL \% $(\mathbf{\uparrow})$ \\[0.05in]
            \toprule
\rownumber 4-Action &
$34.4 (33.6) $  & 
$9.58 (9.52) $
\\
\rownumber 6-Action &
 $30.8 (30.4) $
 & 
 $7.60 (7.28) $
\\
\rownumber 6-Action + Tether &
 $26.6 (27.5) $ & 
 $9.79 (9.23) $
\\
            \bottomrule

        \end{tabular}
    }
    \vspace{5pt}
    \caption{Comparing \textsc{val} scores when providing vertical dimension in GPS. Formatted as 2D (3D). There is little difference in performance.}
    \label{tab:supp_2d3d}
\end{table}

Though we train with the 3D GPS sensor provided by the simulator, all presented results have used a 2D GPS without the vertical dimension at evaluation, to be consistent with Habitat Challenge settings. We compare 2D vs 3D validation scores in~\tableref{tab:supp_2d3d}. We do not see a large change in performance; our agents dos not leverage the vertical dimension much.

\subsection{Additional Auxiliary Task Ablations}
\label{sec:aux_sge}

\begin{table}[t]
    \centering
    \resizebox{0.8\linewidth}{!}{
        \begin{tabular}{@{}lccc@{}}
            & Success \% $(\mathbf{\uparrow})$ & SPL \% $(\mathbf{\uparrow})$ \\[0.05in]
            \toprule
\rownumber 4-Action &
$34.4 $\scriptsize{$\pm 2.0$}
&
$9.58 $\scriptsize{$\pm 0.75$}
\\
\rownumber \: - SGE &
 $20.3 $\scriptsize{$\pm 1.7$}
 & $4.14 $\scriptsize{$\pm 0.47$}
\\
\cdashline{1-4}
\rownumber Aux SGE &
$17.7 $\scriptsize{$\pm 1.6$}
&
$4.03 $\scriptsize{$\pm 0.48$}
\\
\rownumber No Aux &
$30.4 $\scriptsize{$\pm 2.0$} &
$6.56 $\scriptsize{$\pm 0.57$}
 \\
            \bottomrule

        \end{tabular}
    }
    \vspace{5pt}
    \caption{Additional ablations on 4-action base for 1. incorporating SGE as an auxiliary task (Aux SGE) and 2. not providing any auxiliary tasks (No Aux).}
    \label{tab:supp_more_ablations}
\end{table}

We provide 2 additional ablations: 1. removing SGE by turning it into a feature and 2. removing all auxiliary tasks (and using one large RNN). Results are in~\tableref{tab:supp_more_ablations}. 

\xhdr{SGE works better as a feature than as an auxiliary task} (row 2 vs 3). Since SGE is a feature that is easily derived from the semantic input, its large contribution to learning efficiency may feel unintuitive. However, when we encourage SGE by decoding it with an auxiliary task, performance counterintuitively decreases. The auxiliary task is easily learned, i.e. its loss flatlines near 0 quickly, and module weights suggest this content is stored in sparsely (\ie $<10$ peaks in the probe weights). This result may warrant additional investigation to understand what types of priors are best introduced as auxiliary tasks \vs as features.

\xhdr{Auxiliary tasks contribute to performance} (row 1 vs 4). If we remove all auxiliary tasks, performance drops moderately. 

\subsection{Behavioral Analysis Details}
\label{sec:error_full}

\begin{table*}[h]
    \centering
        \begin{tabular}{p{2cm}p{6cm}p{6cm}}
            Name & Description & Rule\\[0.05in]
            \toprule
        Plateau & 
        Repeated collisions against the same piece of debris causes a plateau in coverage. Includes debris which traps agent in spawn. & Agent collides $> 50$ times before visiting 2 new voxels.
        \\
        Loop & 
        Poor exploration due to looping over the same locations or backtracking. & 
        Episode $>250$ steps,
        expected fraction of episode spent in current voxel $>0.15$ and collisions $<50$.
        \\
        Detection &
        Despite positive SGE, the agent does not notice nor successfully navigate to the goal. & Max SGE seen in episode $\in [0.02, 0.1]$.
        \\
        Commitment &
        Sees and approaches the goal but passes it. &
        In first 200 steps, SGE $> 0.1$ and goal distance $<1.0$.
        \\
        Last mile &
        Gets stuck near the goal. &
        On stop, SGE $>0$ and distance is $<1m$.
        \\ 
        Open &
        Explores an open area without any objects. &
        Coverage $> 10$ voxels and $< 10$ collisions.
        \\ 
        Quit &
        Quitting despite lack of obstacles. &
        Apply if above rules do not apply and episode length $< 250$.
        \\ 
        Void &
        The goal is seen through a crack in the mesh, which appears to disturb agent behavior. &
        Qualitative.
        \\ 
        Goal Bug &
        A goal instance has no associated viewpoints where the agent can stop. & Qualitative.
        \\
        Explore & 
         A generic failure to find the goal despite steady exploration. Includes semantic failures \eg going outdoors to find a bed. & The default if no other modes apply.
        \\
        \bottomrule
    \end{tabular}
    \caption{Description of observed failure modes for 6-action agents and associated heuristics.}
    \label{table:failure_desc_full}
\end{table*}

\begin{figure}[t]
    \centering
    \includegraphics[width=0.4\textwidth]{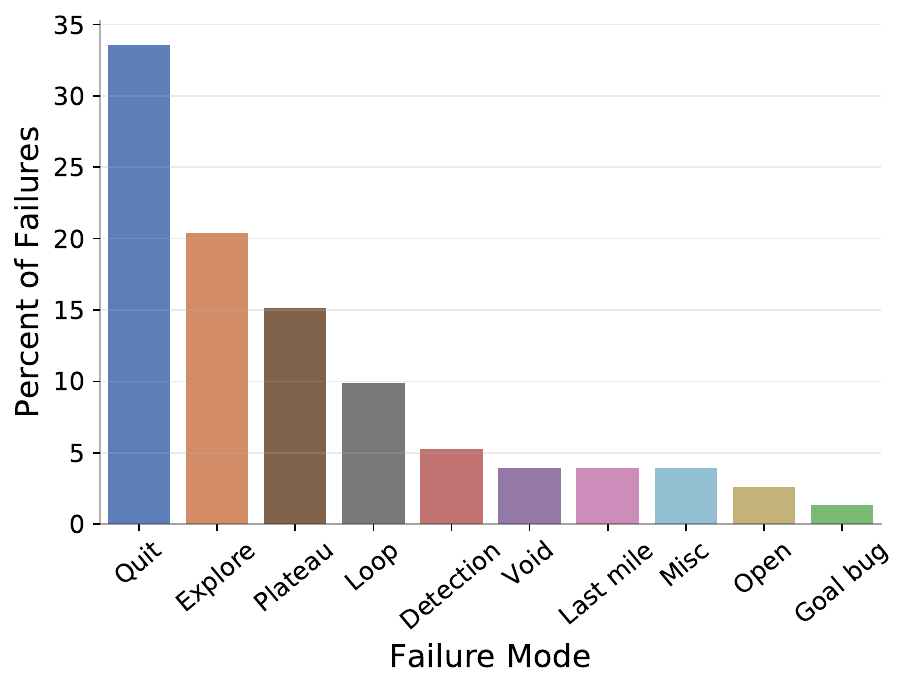}
    \caption{We provide breakdown of tether failure modes.}
    \label{fig:supp_tether_breakdown}
\end{figure}

We provide a failure mode breakdown for the tethered agent (\figref{fig:supp_tether_breakdown}). The tethered agent fails minimally from exploration-reward specific failures (\eg ``Commitment''), but qualitatively fails from a ``Quitting'' mode often, where the agent stops despite reasonable exploration and no goal in sight. We also provide a full list of observed failure modes (\tableref{table:failure_desc_full}), along with quantitative heuristics which is consistent with manual annotations $>80\%$ of the time, to help give a sense of the annotation criteria.

\subsection{Action Entropy to quantify agent instability.}
\label{sec:action_ent}
\begin{figure}[t]
    \centering
    \includegraphics[width=0.4\textwidth]{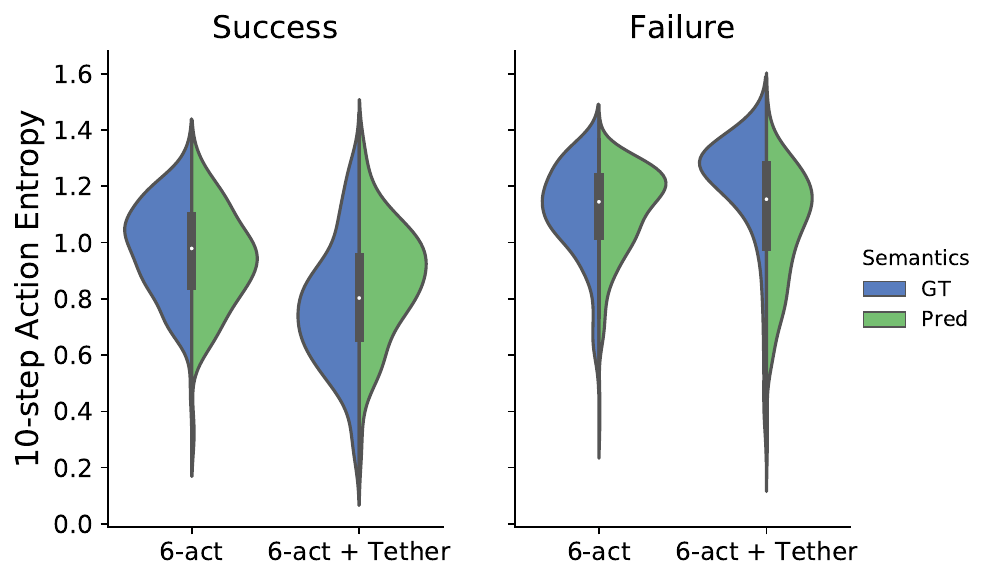}
    \caption{We plot the average action entropy of a rolling window as measured on 300 \textsc{val} episodes, for 2 agents and with and without GT segm. Failures tend to correlate with higher action entropy, more strongly for the tethered agent. Predicted segm. pulls action entropy toward an intermediate value for the tethered agent \ie destabilizes successful episodes, and stabilizes failure episodes. GT and predicted segmentation effects are muted for 6-action agent, perhaps due to its wandering behavior.}
    \label{fig:action_ent}
\end{figure}

Throughout our analysis we have referred to unstable agent behavior. We quantify this instability by measuring action entropy, \ie the entropy of the distribution of actions taken over the course of each episode. Specifically, we measure for each episode the action entropy of a rolling window of 10 steps, averaged across window locations, and show how its distribution differs between agents and episode success or failure in~\figref{fig:action_ent}.

\subsection{Additional Probing Results}
\label{sec:probes_supp}

\begin{figure}[t]
    \centering
    \includegraphics[width=0.4\textwidth]{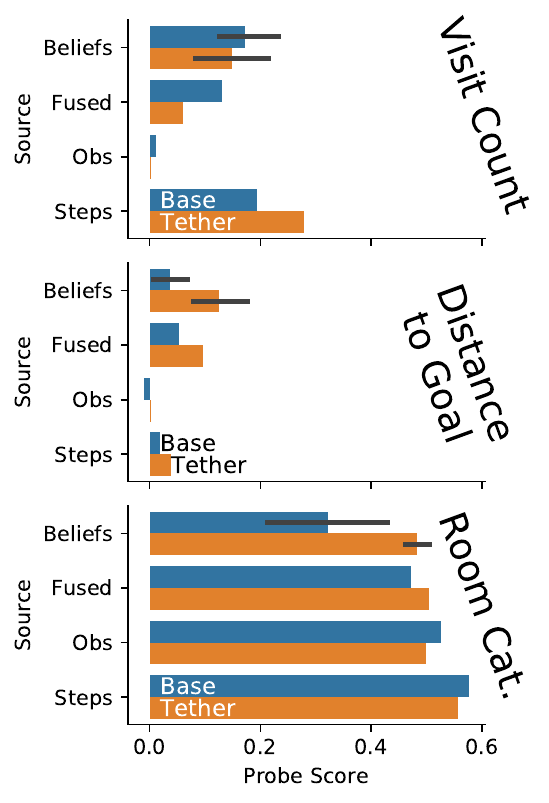}
    \caption{We run the same probing procedure for visit count (number of timesteps spent in current voxel), distance to goal, room category (room category reports classification accuracy instead of $R^2$). Visit count and room category is better matched by timestep than any belief, and distance to goal $R^2$ is at best around 0.1. }
    \label{fig:supp_probes_fig}
\end{figure}

We additionally run probes for time spent in location (as in the ``loop'' failure mode), room ID, and distance to goal, as shown in~\figref{fig:supp_probes_fig}. These features appear negligibly represented across beliefs.

\subsection{Curvature Computation}
\label{sec:curvature}
We use curvature to summarize the stability of a given sequence of representations $x_1 \dots x_t$. We start by calculating normalized displacement vectors $v_i = x_{i+1} - x_i, \hat{v_i} = \frac{v_i}{||v_i||}$. We compute the local discrete curvature as the angle between successive vectors: $c_i = \arccos(\hat{v_i}, \hat{v_{i+1}})$. Then we report the global curvature of the sequence as the mean of these local curvatures. This procedure mirrors~\citep{henaff2019perceptual}. Global curvatures reported in text are averaged over validation episodes.

\subsection{Fixed Point Analysis Methodology}
\label{sec:fp_methods}

Fixed point (FP) analysis of RNNs center around the idea that an RNN's nonlinear dynamics and computation can be understood through its behavior around a set of ``slow (fixed) points.'' These points act as a dynamical skeleton which \eg determines the flow field at other points in hidden space. For example, it has been shown that sentiment analysis RNNs act as line attractors, where the hidden state's position along a line represents the read out sentiment~\citep{maheswaranathan2020recurrent}.

FP analysis begins with an optimization to identify these slow points, \ie hidden states that satisfy $||\text{RNN}(h, \phi) - h||^2 < \epsilon$, for some input $\phi$ and threshold $\epsilon$. For this initial analysis we follow~\citep{maheswaranathan2019reverse} and let $\phi$ be the average input, \ie the average of observation embeddings collected across the 300-episode \textsc{val} subset. Note that this average includes an SGE $> 0$ signal; in case this has large impact on belief dynamics, we verified that fixed point results were qualitatively similar after setting this signal to $0$. We optimize 10K fixed points, by sampling 10K random hidden states experienced in the 300-episode subset and performing gradient descent.

We optimize fixed points for each belief, on each agent; by optimization termination most points have an RNN update norm $< 1\times 10^{-6}$, though we take a subset with norms $< 4\times 10^{-7}$. From this population of slow points we sample 50 points for~\figref{fig:memory_rank}. Then, we compute the Jacobian of the RNN update with respect to the hidden state, and compute the Jacobian's eigendecomposition. The Jacobian will likely have complex eigenvalues; to convert these into a term representing memory, we compute a time constant (as in~\citep{maheswaranathan2019reverse}:
\begin{align}
    \tau(\lambda) = |\frac{1}{\log ||\lambda||}|
\end{align}

\begin{figure}[t]
    \centering
    \includegraphics[width=0.4\textwidth]{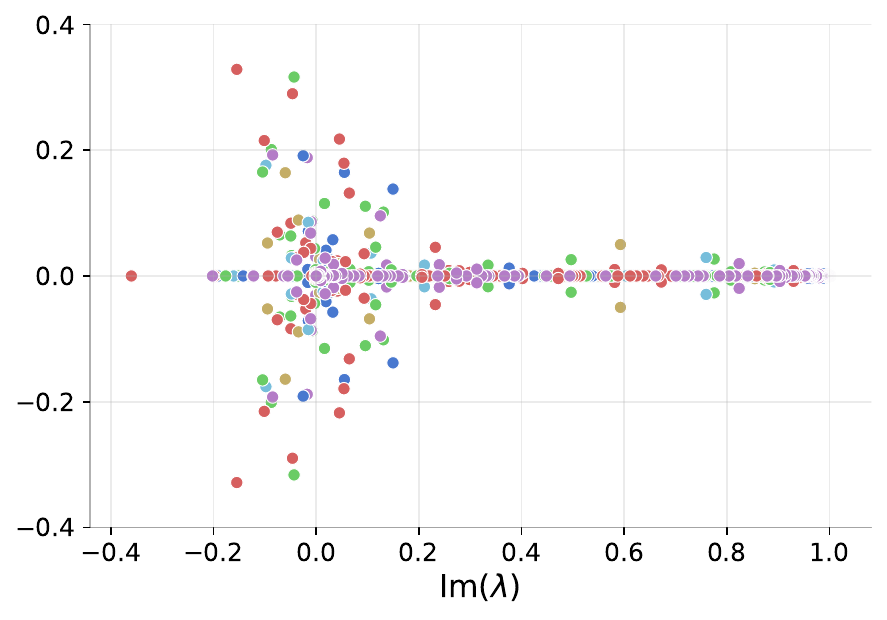}
    \includegraphics[width=0.4\textwidth]{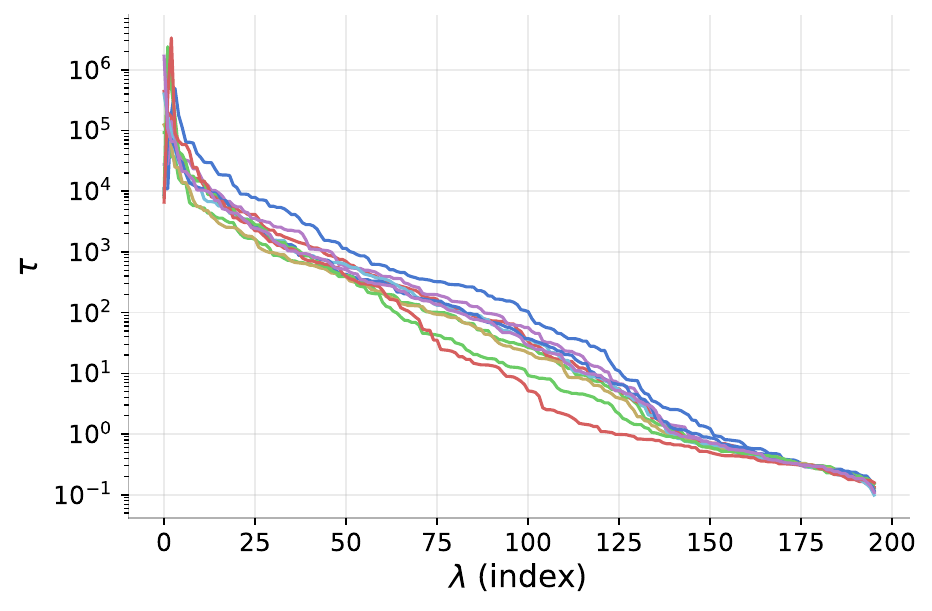}
    \caption{As an example, for the Jacobians of 10 sample fixed points from the PBL belief of the tether agent, we plot Top: their eigenvalues, Bottom: the associated time constants. Different fixed points have different colors.}
    \label{fig:supp_fp_spectrum_ex}
\end{figure}

We show an example eigenvalue spectrum and associated time constants in~\figref{fig:supp_fp_spectrum_ex}. There does not appear to be large variation in spectra across fixed points.

\subsection{High Dimensional Computation in RNNs}
\label{sec:fp_dim}

\begin{figure}[t]
    \centering
    \includegraphics[width=0.4\textwidth]{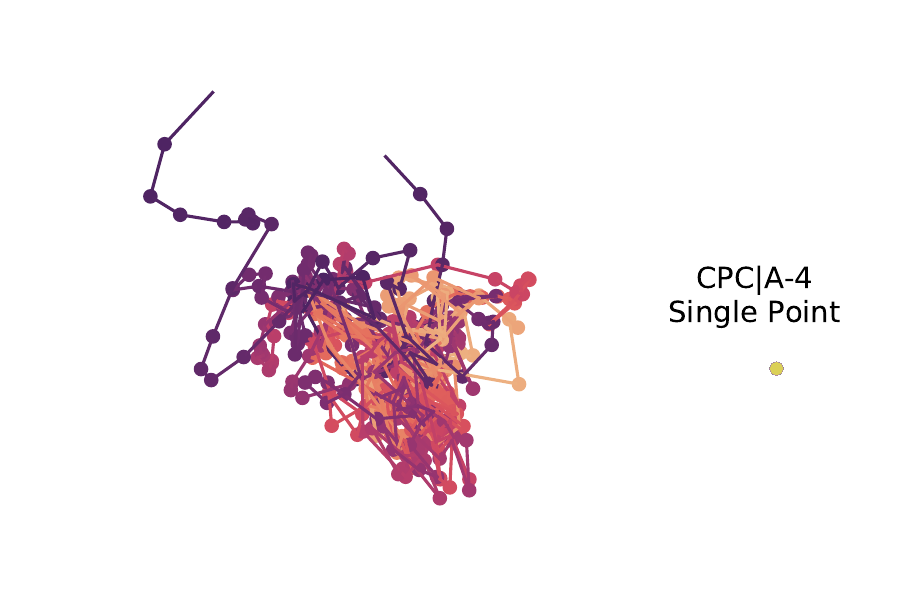}
    \includegraphics[width=0.4\textwidth]{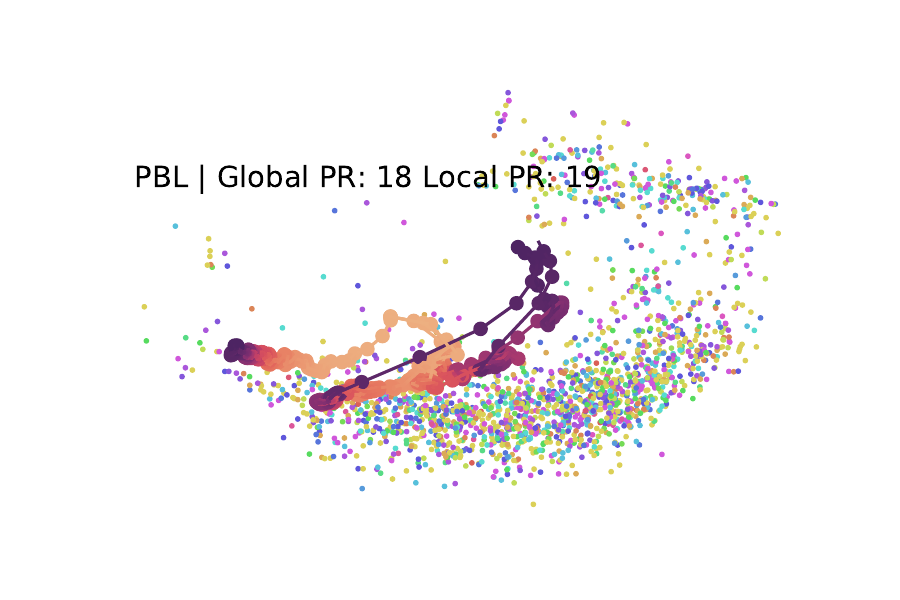}
    \includegraphics[width=0.4\textwidth]{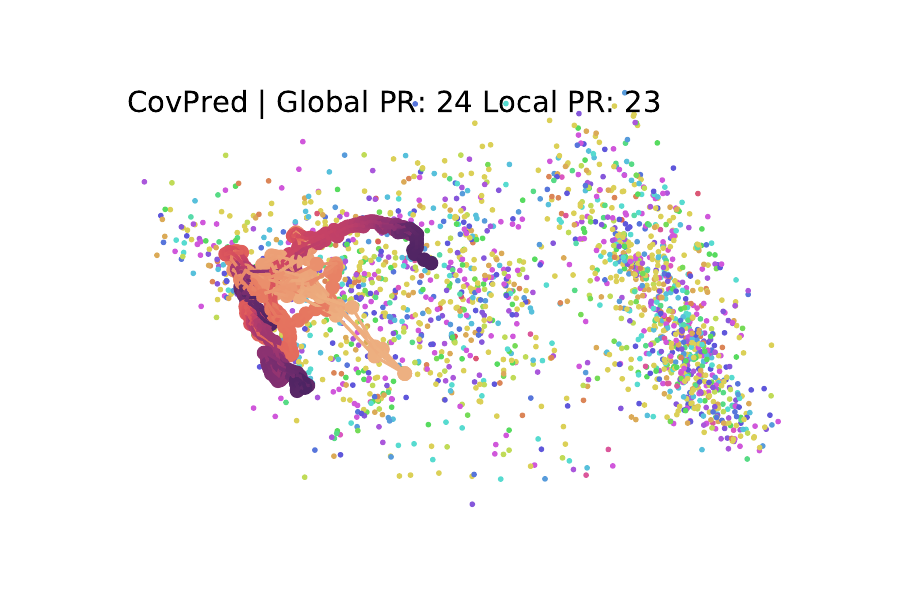}
    \caption{For all beliefs of the tether agent, we plot fixed points (colored by goal category of the state that the fixed point was initialized at) along with sample hidden state trajectories (connected with lines). Plots are in the top-2 PCs of the hidden states. Layouts are different across beliefs, but the top-2 PCs only account for a moderate amount of variance among the fixed points. The RNN correseponding to \cpcat4 only has 1 unique fixed point, but it is not solely an attractor.}
    \label{fig:supp_fp_subspace}
\end{figure}

Prior works find FPs to be arranged in low-dimensional manifolds and are able to directly link RNN memory structure to, \eg, ring attractors and simplicial structures. We find that the RNN FP subspace for \objnav is high-dimensional, as measured \eg by the participation ratio of the fixed point subspace. We believe this is due to 1) \objnav being considerably more complex than previously studied tasks, and 2) unconverged training. This high-dimensionality makes it is difficult to clearly link agent beliefs to any known attractors, \eg inputs project to $>3$ dimensions. Though we observe certain behavioral phenomena that we expect to be reflected in RNN dynamics, \eg the agent will do a near $360\degree$ when it is blocked, we find this difficult to visualize. We show examples fixed point subspaces for 3 beliefs in~\figref{fig:supp_fp_subspace}. These layouts are not qualitatively consistent across agents, nor beliefs, even though memory structure is.

While 20D fixed point subspaces do frustrate current techniques, we could alternately be surprised that the RNN which has 196 dimensions has such a relatively low-dimensional subspace. It would be valuable to both the vision community and beyond to understand how to scale up this technique to a moderately higher number of dimensions.

\subsection{Agent Embedding Visualization}
\label{sec:embeddings}
\begin{figure}[t]
    \centering
    \includegraphics[width=0.4\textwidth]{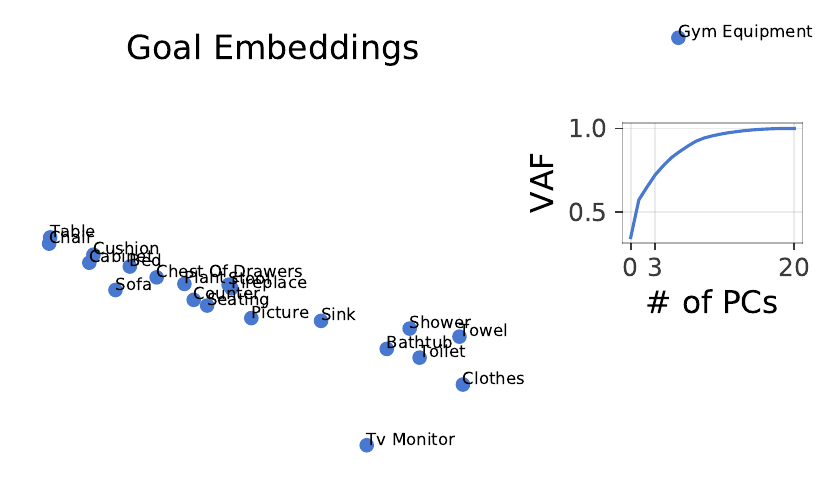}
    \caption{We project goal and semantic channel embeddings into their top-2 PCs. (Inset indicates variance accounted for by PCs, title indicates that participation ratio (approximate dimensionality) of goal is around 10.}
    \label{fig:goal_embed}
\end{figure}

A common way to gain intuition for agent knowledge is by examining agent embedding layers. Our agent incorporates two semantic embedding layers, one which embeds the semantic frame (for the ResNet), and one which embeds the semantic goal; we visualize the goal embedding with PCA in~\figref{fig:goal_embed}. It is difficult to draw falsifiable conclusions about these embeddings. For example, though there are promising clusters of (table, chair, cushion) and (toilet, sink, bathtub, shower) goals, this is also confounded as those same categories tend to have high average success or failure, respectively; \ie instead of room semantics, the embeddings may simply imply goal difficulty. We do not show the semantic channel embedding as there does not appear to be any meaningful arrangement in PC-space.

\subsection{Tethered Policy Updates}
\label{sec:tethered}
We tether a second policy to the acting policy by sharing a base network (we only split policies at the linear actor-critic heads) and training the second policy with a different reward. The first policy is naturally off-policy for the second policy; we thus incorporate V-Trace-like~\citep{espeholt2018impala} off-policy importance weighting terms into the standard on-policy PPO update. Specifically, we 1. replace the tethered policy gradient ratio $r = \frac{\pi^\text{tether}(a|s)}{\pi_\text{old}^\text{tether}(a|s)}$ with $\frac{\pi^\text{tether}(a|s)}{\pi_\text{old}^\text{act}(a|s)}$ and 2. add a clipped importance weight term $c$ to the value function target from 
\[
v_s = V(x_s) + \sum_{t=s}^{s+n-1} \gamma^{t-s}\tau\delta_tV
\]
to 
\[
v_s = V(x_s) + \sum_{t=s}^{s+n-1}\gamma^{t-s}\tau \text{clip}(\frac{\pi^{\text{tether}}(a|s)}{\pi^{\text{act}}(a|s)}, a, b)\delta_tV
\]
With $\delta_tV = r_t + \gamma V_{x_{t+1}} - V_{x_t}$ being a TD for $V$. We set clip terms $a = 0.01, b=1.0$. The overall RL loss (both actor and critic losses) are averaged across policies, \ie equal weighting of acting and tethered policy loss.

\subsection{Training and Auxiliary Task Details}
\label{sec:training_details}

We train our agent via Proximal Policy Optimization
PPO)~\citep{schulman_arxiv17} with Generalized Advantage Estimation
(GAE)~\citep{schulman_iclr16} and using the Adam optimizer~\citep{kingma_iclr15}. Agent parameter counts were all $5-6$ million parameters, excluding parameters in auxiliary modules. This amounts to GRU hidden sizes of 196 (except for the No Aux ablation, which has 512). General hyperparameters follow~\citep{ye2020aux}:
\begin{align}
    \text{Rollout Workers: } n &= 4 \\
    \text{Rollout Length: } t &=128 \\
    \text {PPO Epochs} &= 4 \\  
    \text {PPO Mini-batches} &= 2 \\
    \text {Discount } \gamma &= 0.99 \\
    \text{GAE } \tau &= 0.95 \\
    \text{lr} &= 2.5 \times 10^{-4} \\
    \text{Adam } \epsilon &= 1\times10^{-5} \\
    \text{Gradient Norm Cap} &= 0.5 \\
    \text{PPO Clip} &= 0.2
\end{align}

except PPO Clip factor, which was set to $0.2$ instead of $0.1$, as per recommendations in~\citep{wijmans2020train}.

Our complete loss is:
\begin{align}
    L_\text{total}(\theta_m;\theta_a) & = L_\text{RL}(\theta_m) - \alpha H_{action}(\theta) + L_\text{Aux}(\theta_m;\theta_a)
    \\ 
    L_\text{Aux}(\theta_m;\theta_a) &= \sum_{i=1}^{n_\text{Aux}}\beta^{i}L^i_\text{Aux}(\theta_m;\theta_a^i) - \mu H_{attn}(\theta_m) 
\end{align}

$H_{attn}$ is the entropy of the attention distribution over the different auxiliary tasks. We set $\alpha = 0.01$ for 4-action agents (as in~\citep{ye2020aux}) and $\alpha = 0.0075$ for 6-action agents. We set $\mu = 0.075$, which amounts to belief weighting across to be $\approx$ equal; we find little difference when weighting is more flexible. Auxiliary task loss coefficients $\beta$ were determined such that the loss terms were in the same order of magnitude at initialization.

For extended details, the configurations of our experiments are available in the code release.

\subsection{Rednet Tuning}
\label{sec:rednet}

\begin{table}
\centering
 \resizebox{\linewidth}{!}{
 \begin{tabular}{c c c c c c } 
 \toprule
&  & mIoU & mRecall & mPrecision & Acc \\  
 \midrule
\multirow{2}{*}{\rotatebox{90}{train}} & 40-class & 77.09 & 86.46 & 87.22 & 91.62 \\
& 21-class & 37.13 & 92.25 & 39.23 &  81.96 \\
\midrule
\multirow{2}{*}{\rotatebox{90}{val}} & 40-Class & 24.95 & 36.96 & 39.62 & 61.88 \\ 
& 21-Class & 15.93 & 39.45 & 21.65 & 69.01 \\
 \bottomrule
\end{tabular}
}
\vspace{5pt}
\caption{RedNet \cite{jiang2018rednet} performance on the MP3D dataset after tuning on 40 MP3D categories, or after tuning on 21 goal categories. Note that these numbers include accuracy on \textsc{VOID} class which skew numbers (especially accuracy) upward.}
\label{table:rednet_results}
\end{table}
We use RedNet to predict segmentation from RGBD at test time. We finetune the model (which was pretrained on SUNRGBD) with 100K randomly sampled front-facing views rendered in the Habitat simulator (16K validation views). This procedure is the one used in~\cite{cartillier2020semantic}. We initially trained the model to segment all 40 MP3D categories; we present its accuracies in~\tableref{table:rednet_results}. Unlike our agent, this RedNet is greatly overfit. Since our agent did not appear to greatly leverage semantic cues during exploration (through~\secref{sec:error_anlaysis},~\secref{sec:embeddings}), we reduced the complexity of the RedNet task by asking it to only segment the 21 goal categories (other categories were cast to \textsc{void}). This improves performance of segmentation on the goal categories, and resulted in slight improvement in validation scores after 10M frames of agent tuning, so we used it in the main text. 

\subsection{Negative Results}
In the course of our experiments, we found that:
\begin{itemize}
    \item Adding a curiosity module (ICM~\citep{pathak_icml17}) as non-episodic intrinsic reward failed to change performance significantly.
    \item Forcing agent recall (to fix the ``loop'' failure mode) with an Action Recall auxiliary task failed to change agent performance. This task was implemented as ADP with a negative horizon.
    \item Controlling the agent's sense of time by projecting beliefs out of the probed time dimension was unstable. We could not \eg solve ``Commitment'' failure modes by setting the time variable to near end-episode without degrading performance.
\end{itemize}

\subsection{RedNet vs GT with CI}
\begin{table*}[t]
    \centering
    \resizebox{1.8\columnwidth}{!}{
        \begin{tabular}{@{}lccccc@{}}
            & \multicolumn{2}{c}{Success \% $(\mathbf{\uparrow})$} & & \multicolumn{2}{c}{SPL \% $(\mathbf{\uparrow})$ } \\
            \cline{2-3}
            \cline{5-6}
            &
            \textsc{train} & \textsc{val}
            & &
            \textsc{train} & \textsc{val} \\[0.03in]
            \toprule
        \rownumber 4-Act &
        $50.3 $\scriptsize{$\pm 5.7$}
 (
 $36 $\scriptsize{$\pm 5.5$}
 ) 
 & 
        $43.3 $\scriptsize{$\pm 5.6$}

        (
        \boldsymbol{
            $34.4 $
         }\scriptsize{\boldsymbol
             {$\pm 2.0$}
         }  
        ) %
        & &
        $18.1 $\scriptsize{$\pm 2.7$}
        (
        $12.4 $\scriptsize{$\pm 2.3$}
        ) & 
        $12.3 $\scriptsize{$\pm 2.1$}
        (
        
        \boldsymbol{
            $9.58 $
         }\scriptsize{\boldsymbol
             {$\pm 0.75$}
         }
        )
        \\
        \rownumber 6-Act &
        $56 $\scriptsize{$\pm 5.6$}
        (
        $21.7 $\scriptsize{$\pm 4.7$}
        ) & 
        \boldsymbol{
            $58.0 $
         }\scriptsize{\boldsymbol
             {$\pm 5.6$}
         }
        (
        $30.8 $\scriptsize{$\pm 1.9$}
        ) %
        & &
        $21.5 $\scriptsize{$\pm 2.9$}

        (
        $8.24 $\scriptsize{$\pm 2$}
        ) & $16.9 $\scriptsize{$\pm 2.3$}
        (
        $7.60 $\scriptsize{$\pm 0.64$}
        )
        \\
        \rownumber 6-Act + Tether &
        $54 $\scriptsize{$\pm 5.7$}
        (
        $27.3 $\scriptsize{$\pm 5.1$}

        ) & 
        $48.7 $\scriptsize{$\pm 5.7$}

        (
        $26.6 $\scriptsize{$\pm 1.9$}
        )
        & &
        $27.9 $\scriptsize{$\pm 3.5$}
 ($11.5 $\scriptsize{$\pm 2.5$}
) &        
        \boldsymbol{
            $19.1 $
         }\scriptsize{\boldsymbol
             {$\pm 2.7$}
         }

        (
        \boldsymbol{
            $9.79 $
         }\scriptsize{\boldsymbol
             {$\pm 0.82$}
         }
        ) %
        \\
    \bottomrule
    \end{tabular}}
    \vspace{5pt}
    \caption{Performance on a 300-episode subset of \textsc{train} and \textsc{val}
        splits, reported as ``with GT segmentation (with RedNet segmentation)''. Reproducing~\tableref{tab:gt_segm} with 95\% CIs included. Bold values are significantly better ($p < 0.05$) than non-bold values in a paired t-test on 300 episodes.
        Note that the \textsc{val} performance with RedNet is reported on the full split, taken from the primary experiments.
    }
    \label{tab:supp_gt_segm_full}
\end{table*}
We present a full version of~\tableref{tab:gt_segm} with 95\% CI estimates in~\tableref{tab:supp_gt_segm_full}.

\end{document}